\documentclass{article} 
\usepackage[final]{colm2026_conference}

\usepackage{microtype}
\usepackage{hyperref}
\usepackage{url}
\usepackage{booktabs}
\usepackage{graphicx}
\usepackage{amsmath}
\usepackage{booktabs}
\usepackage{siunitx}
\usepackage{colortbl}
\usepackage{subcaption}
\usepackage{multirow}
\usepackage{etoc}
\usepackage[ruled,vlined]{algorithm2e}
\usepackage[most]{tcolorbox}
\usepackage{float}
\usepackage{mdframed}
\usepackage{hyperref}

\usepackage{lineno}

\definecolor{darkblue}{rgb}{0, 0, 0.5}
\hypersetup{colorlinks=true, citecolor=darkblue, linkcolor=darkblue, urlcolor=darkblue}

\title{\textsc{\ours}: Inference-Time Weak-to-Strong Generalization from Small Language Model Failure Modes}

\author{
Yufan Wu$^{1}$,
Yinghui He$^{2}$,
Zhengyi Hu$^{1}$,
Lang Wei$^{1}$,
Ruichen Li$^{1}$,
Qifan Yang$^{1}$,
Ting Zhu$^{1}$ \\
$^{1}$The Ohio State University \quad $^{2}$Princeton University \\
\texttt{\{wu.6545, zhu.3445\}@osu.edu}
}

\newcommand{\ours}{CritICL}
\newcommand{\dataset}{CritBank}

\newtcolorbox{promptbox}[1]{
  colback=gray!5,
  colframe=gray!75!black,
  fonttitle=\bfseries,
  title=#1,
  boxrule=0.8pt,
  arc=3pt,
  left=6pt,
  right=6pt,
  top=6pt,
  bottom=6pt
}
\newmdenv[
  backgroundcolor=gray!5,
  linecolor=gray!60,
  linewidth=0.8pt,
  roundcorner=3pt,
  skipabove=8pt,
  skipbelow=8pt,
  innerleftmargin=6pt,
  innerrightmargin=6pt,
  innertopmargin=6pt,
  innerbottommargin=6pt
]{graybox}

\begin{document}

\ifcolmsubmission
\linenumbers
\fi

\maketitle

\begin{abstract}
Recent advances in inference-time scaling have significantly improved the reasoning performance of large language models (LLMs). However, these methods typically rely on repeated generation or external verification. To address this limitation, we introduce \ours{}, a novel inference-time framework that improves reasoning while maintaining high efficiency.

Our key insight is that LLM failure modes exhibit structured patterns across model scales within the same family. Instead of treating failures as undesirable outputs, \ours{} leverages them as a source of guidance. Specifically, we utilize failure modes derived from weaker models and incorporate them into inference through critique-based in-context examples. We propose two variants: \ours-dynamic, which adaptively predicts input-specific failure modes and retrieves critiques, and \ours-static, which uses a global failure mode profile to provide stable guidance.

Experimental results show that \ours{} consistently outperforms standard in-context learning and achieves performance competitive with or superior to test-time scaling methods, while requiring significantly fewer generations and lower token cost.

\begin{center}
{\fontsize{11pt}{11pt} \selectfont \raisebox{-0.06em}{\includegraphics[height=1em]{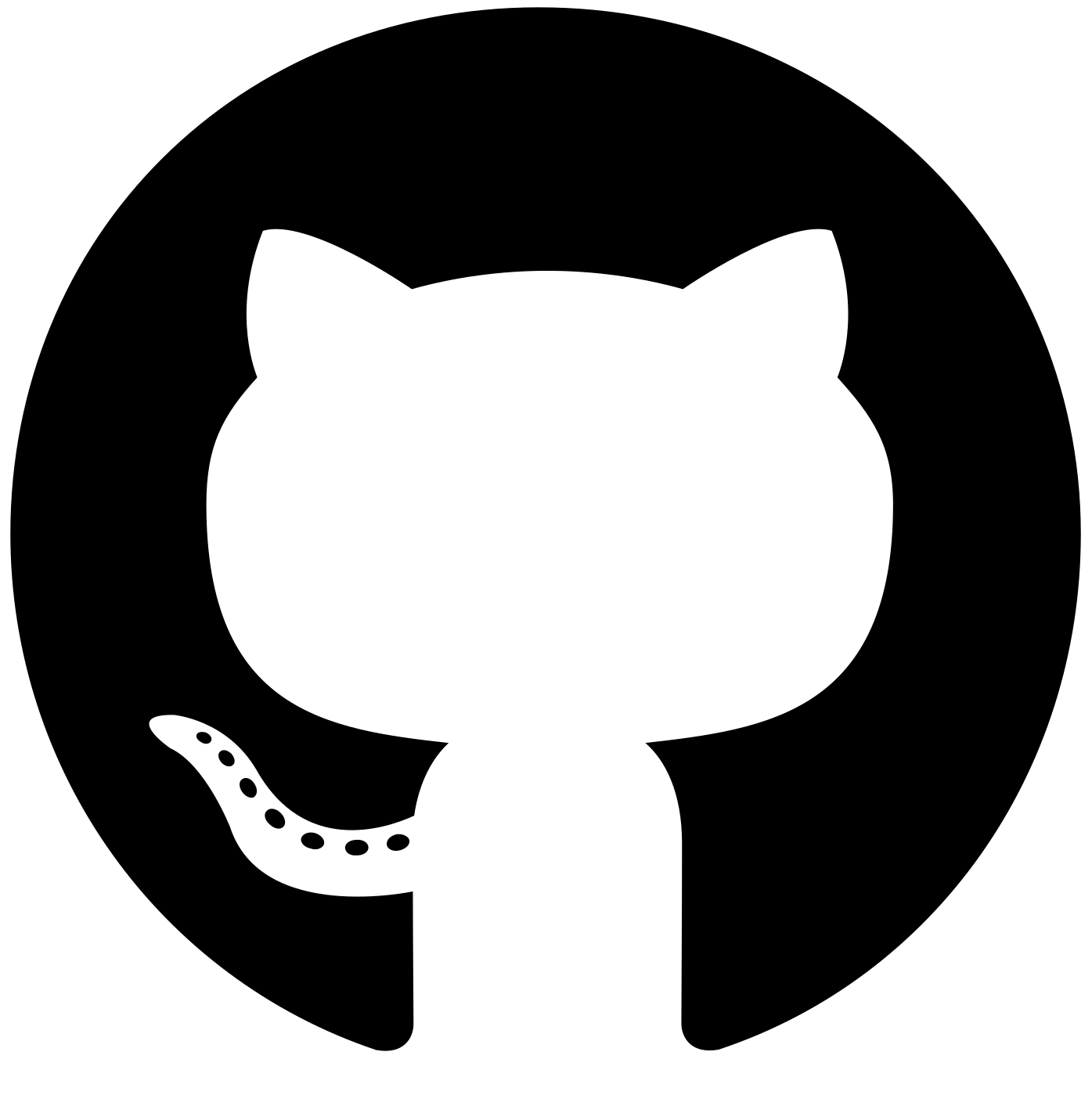}}} \href{https://github.com/umwyf/CRITICL}{https://github.com/umwyf/CRITICL} 
\end{center}
\end{abstract}

\section{Introduction}







Inference-time scaling has emerged as a promising way to improve the reasoning performance of large language models (LLMs), with prior work showing gains through repeated sampling \citep{wang2022self}, iterative self-refinement \citep{madaan2023selfrefine, shinn2023reflexion}, and external verification \citep{zheng2023judging, huang2024llmjudge}. However, these improvements often come at a substantial inference cost, often requiring multiple generations either from the model itself or from a stronger model.

A more recent line of work explores whether weaker models can provide useful inference-time guidance to stronger ones \citep{ding2026w2saligntreeweaktostronginferencetimealignment}. Such approaches typically rely on weak models to produce online supervision or intermediate guidance for each new input, which still introduces additional inference overhead and may depend on the quality of the weaker model’s direct outputs. More importantly, they do not fully exploit a potentially richer source of transferable signal: the systematic ways in which models fail.

In our work, we draw inspiration from a fundamental property of LLM reasoning: model errors are often not arbitrary, but structured and predictable \citep{didolkar2024metacognitive}. Instead of viewing weaker model failures as merely undesirable outputs, we treat them as a source of structured \emph{failure modes}. In particular, our experiments in Section~\ref{dis:why_works} show that the relative distribution of stronger models' (Qwen, 72B) failure modes remains highly consistent with that of much weaker models (Qwen, 1.5B) from the same family. This suggests that weak and strong models share a common failure structure, even when their capabilities differ substantially.



This observation motivates a key question:

\begin{quote}
Can we leverage the structured failure modes of weaker models to improve stronger models at inference time, with minimal additional inference cost?
\end{quote}




\begin{figure}[t]
    \centering
    \includegraphics[width=1.0\linewidth]{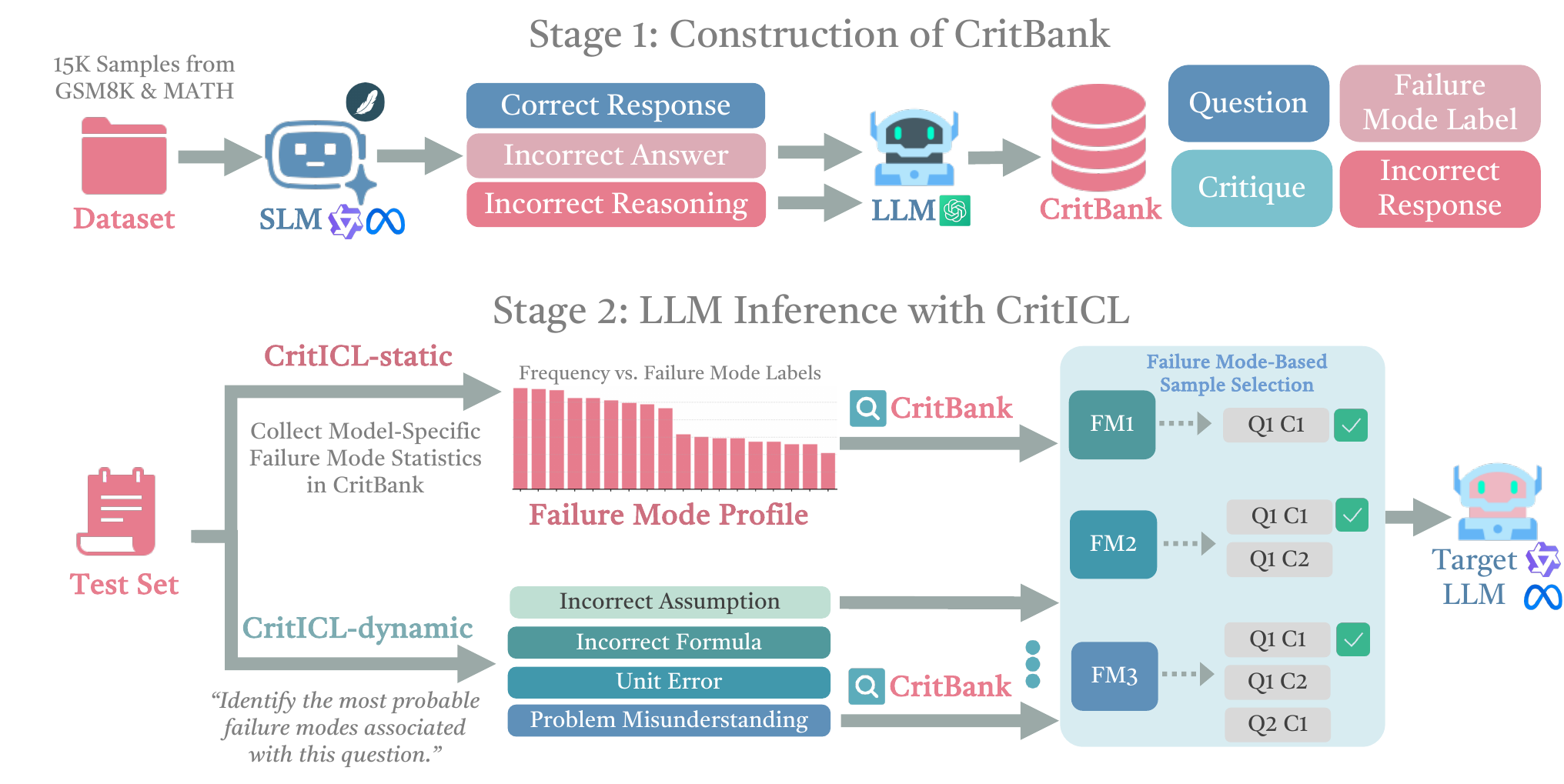}
    \vspace{-10pt}
    \caption{\ours{} is a two-stage, inference-time W2SG method. \textbf{Stage 1:} Construct \dataset{}. We use SLMs to generate responses and reasoning on a dataset. Then we leverage a frontier LLM to produce critiques and assigns failure mode labels. \textbf{Stage 2:} Perform LLM inference with CritICL. For a given query, \ours-static constructs model-specific failure mode profiles from \dataset{}, while \ours-dynamic identifies failure modes relevant to the query. We then performs failure mode-based in-context sample selection from \dataset{} to provide in-context examples for target LLMs.}
    \label{fig:design}
    \vspace{-10pt}
\end{figure}

We introduce \textbf{CritICL}, a framework that improves LLM reasoning by transferring structured failure modes from weaker models to stronger ones at inference time. Our method is based on the intuition that mistakes made by weaker models encode useful information about recurring reasoning pitfalls, and that this information can be reused as actionable guidance for stronger models.



We first construct \textbf{\emph{CritBank}}, a structured dataset of failure-aware critiques derived from weaker models, where each entry contains a question, an incorrect response, failure mode labels, and a natural language critique. Aggregated across multiple small models and tasks, CritBank captures failure modes shared across model scales. We then introduce two inference-time variants: \textbf{\emph{CritICL-dynamic}}, which adaptively predicts likely failure modes for each model input and retrieves relevant critiques; and \textbf{\emph{CritICL-static}}, which uses a model-family-specific \emph{failure mode profile} to retrieve critiques associated with dominant failure modes. Both variants leverage failure-aware guidance collected offline from weaker models.



We evaluate CritICL on a range of mathematical reasoning benchmarks, including GSM8K \citep{cobbe2021training}, MATH \citep{hendrycks2021measuring}, AMC \citep{amc2023} and AIME \citep{aime24, aime25}. 
Experimental results show that CritICL consistently outperforms standard in-context learning and matches or exceeds test-time scaling methods, while requiring significantly fewer generations and lower token cost.

Further analysis demonstrates that CritICL generalizes beyond mathematical reasoning and can be effectively extended to other domains. These results establish CritICL as an efficient method for improving reasoning at inference time by leveraging failure mode information from weaker models.




\section{Design of \dataset{} and \ours}


Our method builds on the observation that models within the same family exhibit similar distributions of failure modes. This shared structure enables a form of weak-to-strong generalization: failure modes identified from weaker models can effectively transfer to guide stronger models. Our empirical analysis in Section~\ref{dis:why_works} supports this observation and demonstrates that failure modes learned from weaker models remain highly informative for improving the reasoning of more capable models.

\subsection{Construction of \dataset{}}

We now describe the construction of \dataset{}, a structured dataset of questions, incorrect responses, failure mode labels, and critiques (see Figure~\ref{fig:design}).

\paragraph{Response Generation.}
Let $\mathcal{Q}$ denote the set of input questions, and let $M$ be a set of small language models from the same model family. For each $(q,m) \in \mathcal{Q} \times M$, we prompt model $m$ with question $q$ using chain-of-thought (CoT) prompting to generate five responses:
\[
R(q,m) = \{ r_{q,m}^{(i)} \}_{i=1}^5,
\]
where each $r_{q,m}^{(i)}$ contains both intermediate reasoning steps and a final answer. The full collection of responses is
\[
R(\mathcal{Q}, M) = \bigcup_{q \in \mathcal{Q},\, m \in M} R(q,m).
\]
We define a correctness function $\phi(q, r) \in \{0,1\}$ that indicates whether $r$ correctly solves $q$. Our focus is on incorrect responses, i.e., those with $\phi(q, r) = 0$, as they expose the underlying reasoning failures of the model. 

For each $(q,m)$, we partition the response set $R(q,m)$ into correct and incorrect subsets:
\[
R(q,m) = R_{\text{correct}}(q,m) \cup R_{\text{incorrect}}(q,m), \quad
R_{\text{correct}}(q,m) \cap R_{\text{incorrect}}(q,m) = \emptyset,
\]
where $R_{\text{correct}}(q,m) = \{ r \in R(q,m) \mid \phi(q,r) = 1 \}$ and 
$R_{\text{incorrect}}(q,m) = \{ r \in R(q,m) \mid \phi(q,r) = 0 \}$.

\paragraph{Failure Mode and Critique Generation}
For each incorrect response $r \in R_{\text{incorrect}}(q,m)$, we leverage a frontier LLM to generate up to five candidate failure mode labels, each capturing a potential failure mode. As these labels may be noisy or semantically redundant, we apply a clustering procedure to group similar labels and extract a set of representative failure modes. Specifically, we adopt the clustering approach proposed by \citet{didolkar2024metacognitive}. In addition, we generate a natural language critique for each $(q, r)$ pair to provide fine-grained feedback on the reasoning process. Unless otherwise specified, all failure mode labels and critiques are generated using gpt-4o-mini \citep{achiam2023gpt}. Prompt templates used in this process are provided in Appendix~\ref{app:prompt_temp}.

\paragraph{Final Dataset.}
After collecting all samples in \dataset{}, we define two mapping functions over $(q,r)$ pairs. The critique function $\mathcal{C}$ maps each $(q,r)$ pair to a structured critique, while the labeling function $\mathcal{L}$ is a set-valued function that assigns each $(q,r)$ pair a subset of failure modes.

The final dataset is defined as
\[
\dataset{}(\mathcal{Q}, M) = \left\{ (q, r, l, \mathcal{C}(q,r)) \;\middle|\; q \in \mathcal{Q},\ m \in M,\ r \in R_{\text{incorrect}}(q,m),\ l \in \mathcal{L}(q,r) \right\}.
\]

Based on these mappings, we further define the inverse mapping of $\mathcal{L}$ to retrieve all associated tuples for a given failure mode label $l$:
\[
\mathcal{L}^{-1}(l) = \left\{ (q, r) \;\middle|\; l \in \mathcal{L}(q,r) \right\},
\]
and the corresponding set of tuples with critiques:
\[
\left\{ (q, r, \mathcal{C}(q,r)) \;\middle|\; (q,r) \in \mathcal{L}^{-1}(l) \right\}.
\]

This mapping is leveraged in \ours{} to identify and select the most informative in-context samples.

\subsection{\ours{}-dynamic and \ours{}-static}

Given a new question $q' \notin Q$, our goal is to retrieve and utilize informative examples from \dataset{} to construct effective prompts for the target large language model at inference time. We propose two strategies that differ in how they select and incorporate critiques.


\paragraph{\ours-dynamic}
Our first approach performs input-dependent critique selection. 
Given a query $q'$, we prompt the target model to predict a small set of likely failure mode labels (up to five). 
Conditioned on these predictions, we retrieve relevant examples from \dataset{} via a Failure Mode-Based Sample Selection procedure. Unless otherwise specified, we retrieve at most five examples in our experiments.
The retrieved examples are then incorporated into the prompt  and provide targeted critiques that steer the model away from likely mistakes. 

\paragraph{\ours-static}
Our second approach is model-family-aware and input-agnostic. 
We construct a global failure mode profile by aggregating the failure mode distributions of weaker models within the same family, which are also used to build \dataset{}. 
This profile identifies dominant and persistent failure modes shared across the family. 
We then retrieve corresponding critiques from \dataset{} using the same Failure Mode-Based Sample Selection procedure and incorporate them into the prompt.

Further details of both methods and the Failure Mode-Based Sample Selection procedure are provided in Appendix~\ref{app:details}.
\section{Experiment}

\subsection{Experimental Settings}

\paragraph{Dataset.}
We evaluate our method on two widely used mathematical reasoning benchmarks: GSM8K (7.4k training samples and 1.3k test samples) \citep{cobbe2021training} and MATH (7.5k training samples and 5k test samples) \citep{hendrycks2021measuring}. To construct \dataset{}, we sample all the instances from the training split of each dataset, resulting in a combined set of 15k questions. These samples are used to elicit responses from small-scale language models, which serve as the basis for generating failure mode labels and critiques. For evaluation, we report results on the full test sets of GSM8K and MATH. In addition, to assess out-of-distribution generalization, we further evaluate on three competition-level benchmarks: AMC23 \citep{amc2023}, AIME24 \citep{aime24}, and AIME25 \citep{aime25}.

\paragraph{Model Settings.}
\label{sec:model_set}
We evaluate our method on two model families: Qwen and Llama. For the Qwen family, we construct \dataset{} using responses generated by weaker instruction-tuned models, including Qwen2.5-1.5B-Instruct, Qwen2.5-3B-Instruct, and Qwen2.5-7B-Instruct \citep{qwen2.5}. We then evaluate performance on larger models within the same family, namely Qwen2.5-32B-Instruct and Qwen2.5-72B-Instruct. For the Llama family, we build \dataset{} using responses from Llama-3.2-1B-Instruct, Llama-3.2-3B-Instruct, and Llama-3.1-8B. We evaluate on Llama-3.1-70B-Instruct \citep{grattafiori2024Llama}. For all experiments, we adopt greedy decoding with a generation temperature of 0.0 to ensure deterministic outputs.

\paragraph{Baselines.}
We compare \ours{} against a diverse set of baselines, including standard in-context learning methods and test-time scaling approaches.

\begin{itemize}

    \item \textbf{Zero-shot}: We evaluate the base model without any in-context examples.

    \item \textbf{Few-shot (random exemplars), 1/3/5-shot}: We randomly select 1, 3, or 5 correct examples (question–answer pairs) from the training sets of GSM8K and MATH.

    \item \textbf{Few-shot (fixed exemplars), 1/3/5-shot}: We use a fixed set of 1, 3, or 5 selected correct examples.

    \item \textbf{Self-consistency (consistency@3/5/7)} \citep{wang2022self}: We sample multiple reasoning paths (3, 5, or 7 generations at temperature 1.0) and select the final answer via majority voting. Unless otherwise specified, all experiments on test-time scaling method are conducted under the 5-shot setting.

    \item \textbf{Self-reflection} \citep{madaan2023selfrefine,shinn2023reflexion}: We prompt the model to iteratively critique and refine its own outputs using self-generated feedback.

    \item \textbf{LLM-as-a-judge} \citep{zheng2023judging,huang2024llmjudge}: We use a strong language model to evaluate five candidate responses and select the final prediction based on its judgments. In our experiments, we use GPT-4o-mini to judge candidate responses.

\end{itemize}


\begin{table}[t]
\centering
\small
\fontsize{8pt}{10pt}\selectfont
\setlength{\tabcolsep}{4pt}
\renewcommand{\arraystretch}{1}

\begin{subtable}{\linewidth}
\centering
\caption{\textbf{Target Model: Qwen2.5-32B-Instruct}}
\begin{tabular}{lcccccccc}
\toprule
& \multicolumn{3}{c}{\textbf{In-distribution (ID)}} 
& \multicolumn{4}{c}{\textbf{Out-of-distribution (OOD)}} 
& \textbf{Overall} \\
\cmidrule(lr){2-4} \cmidrule(lr){5-8}
\textbf{Method}
& GSM8K & MATH & Avg. 
& AMC23 & AIME24 & AIME25 & Avg. 
& Avg. \\
\midrule
\rowcolor[gray]{0.9}
\multicolumn{9}{l}{\textbf{Standard ICL}} \\

Zero-shot  & 82.4 & 42.1 & 62.3 & 14.8 & 9.6 & 8.7 & 11.0 & 36.6 \\
1-shot (Rand.) & 85.6 & 46.3 & 66.0 & 16.5 & 11.2 & 10.1 & 12.6 & 39.3 \\
3-shot (Rand.) & 88.2 & 50.8 & 69.5 & 18.9 & 13.5 & 12.3 & 14.9 & 42.2 \\
5-shot (Rand.) & 90.3 & 54.6 & 72.5 & 22.5 & 15.8 & 14.2 & 17.5 & 45.0 \\

1-shot (Fixed) & 86.8 & 47.5 & 67.2 & 17.2 & 11.9 & 10.7 & 13.3 & 40.3 \\
3-shot (Fixed) & 89.4 & 52.1 & 70.8 & 20.1 & 14.3 & 13.0 & 15.8 & 43.3 \\
5-shot (Fixed) & 91.2 & 55.3 & 73.3 & 23.8 & 16.9 & 15.3 & 18.7 & 46.0 \\

\rowcolor[gray]{0.9}
\multicolumn{9}{l}{\textbf{Test-Time Scaling}} \\

Consistency@3 & 92.1 & 56.9 & 74.5 & 25.2 & 17.9 & 16.3 & 19.8 & 47.9 \\
Consistency@5 & 93.0 & 58.2 & 75.6 & 26.1 & 19.0 & 17.3 & 20.8 & 48.9 \\
Consistency@7 & 93.3 & 58.6 & 76.0 & \textbf{26.9} & 19.4 & 17.8 & \textbf{21.4} & 49.5 \\
Self-Reflection & 92.5 & 57.8 & 75.2 & 25.8 & 18.5 & 16.9 & 20.4 & 48.4 \\
LLM-as-Judge & 92.9 & 58.8 & 75.9 & 26.4 & 19.3 & 17.7 & 21.1 & 49.0 \\

\midrule
\rowcolor[gray]{0.9}
\multicolumn{9}{l}{\textbf{\ours{}(ours)}} \\

\ours{}-dynamic & 93.0 & 58.6 & 75.8 & 26.7 & 19.2 & 17.5 & 21.1 & 49.1 \\
\ours{}-static & \textbf{93.6} & \textbf{59.2} & \textbf{76.4} & 26.6 & \textbf{19.5} & \textbf{17.9} & 21.3 & \textbf{49.8} \\

\bottomrule
\end{tabular}
\end{subtable}

\vspace{0.5em}

\begin{subtable}{\linewidth}
\centering
\caption{\textbf{Target Model: Qwen2.5-72B-Instruct}}
\begin{tabular}{lcccccccc}
\toprule
& \multicolumn{3}{c}{\textbf{In-distribution (ID)}} 
& \multicolumn{4}{c}{\textbf{Out-of-distribution (OOD)}} 
& \textbf{Overall} \\
\cmidrule(lr){2-4} \cmidrule(lr){5-8}
\textbf{Method}
& GSM8K & MATH & Avg. 
& AMC23 & AIME24 & AIME25 & Avg. 
& Avg. \\
\midrule
\rowcolor[gray]{0.9}
\multicolumn{9}{l}{\textbf{Standard ICL}} \\

Zero-shot  & 88.5 & 63.2 & 75.9 & 20.5 & 14.2 & 12.8 & 15.8 & 45.8 \\
1-shot (Rand.) & 90.4 & 68.1 & 79.3 & 22.3 & 15.9 & 14.4 & 17.5 & 48.4 \\
3-shot (Rand.) & 92.1 & 73.5 & 82.8 & 25.6 & 18.6 & 16.9 & 20.4 & 51.6 \\
5-shot (Rand.) & 93.2 & 79.2 & 86.2 & 30.5 & 21.7 & 19.8 & 24.0 & 55.1 \\

1-shot (Fixed) & 91.2 & 70.0 & 80.6 & 23.1 & 16.5 & 15.0 & 18.2 & 49.4 \\
3-shot (Fixed) & 92.9 & 75.4 & 84.2 & 27.2 & 19.8 & 18.0 & 21.7 & 52.9 \\
5-shot (Fixed) & 93.8 & 80.5 & 87.2 & 32.0 & 23.0 & 21.0 & 25.3 & 56.3 \\

\rowcolor[gray]{0.9}
\multicolumn{9}{l}{\textbf{Test-Time Scaling}} \\

Consistency@3 & 94.1 & 81.6 & 87.9 & 33.5 & 24.6 & 22.5 & 26.9 & 57.4 \\
Consistency@5 & 95.0 & 83.2 & 89.1 & \textbf{35.8} & 26.3 & 24.5 & \textbf{28.9} & 59.0 \\
Consistency@7 & 94.0 & 81.0 & 87.5 & 33.1 & 24.3 & 22.2 & 26.5 & 57.0 \\
Self-Reflection & 94.6 & 82.5 & 88.6 & 34.4 & 25.4 & 23.3 & 27.7 & 58.2 \\
LLM-as-Judge & 94.8 & 82.9 & 88.9 & 34.8 & 25.7 & 23.6 & 28.0 & 58.5 \\

\midrule
\rowcolor[gray]{0.9}
\multicolumn{9}{l}{\textbf{\ours{}(ours)}} \\

\ours{}-dynamic & 95.1 & 83.3 & 89.2 & 35.6 & 26.2 & 24.2 & 28.7 & 58.7 \\
\ours{}-static & \textbf{95.4} & \textbf{84.0} & \textbf{89.7} & 35.4 & \textbf{26.5} & \textbf{24.6} & 28.8 & \textbf{59.2} \\

\bottomrule
\end{tabular}
\end{subtable}

\caption{Performance comparison on Qwen family models. \ours-dynamic and \ours-static consistently outperform baselines, achieving improvements of up to 12.9\% and 13.4\%, respectively. Compared to test-time scaling methods, both approaches achieve competitive accuracy without requiring extensive repeated LLM inference. The evaluated models are Qwen2.5-32B-Instruct and Qwen2.5-72B-Instruct, while \dataset{} is constructed using Qwen2.5-1.5B-Instruct, Qwen2.5-3B-Instruct, and Qwen2.5-7B-Instruct. All results are reported as Pass@1 accuracy unless otherwise specified.}
\label{tab:qwen_results}
\end{table}

\subsection{Performance of \ours-dynamic and \ours-static}

Table~\ref{tab:qwen_results} presents the performance of \ours-dynamic and \ours-static on Qwen family models. Overall, both methods consistently outperform standard in-context learning baselines and achieve competitive performance compared to test-time scaling approaches. Due to space constraints, we present additional results on the Llama family in Appendix~\ref{app:add_exp_llama}.

On Qwen2.5-32B-Instruct, \ours-static achieves the best overall performance with 49.8\% Pass@1 accuracy, surpassing the strongest test-time scaling baseline (Consistency@7) by 0.3 points while avoiding repeated inference. \ours-dynamic also performs competitively, matching or exceeding most baselines across both in-distribution (ID) and out-of-distribution (OOD) benchmarks. 

 On Qwen2.5-72B-Instruct, similar trends are observed. \ours-static achieves the highest overall accuracy of 59.2\%, outperforming all baselines, including test-time scaling methods such as Consistency@5 (59.0\%). \ours-dynamic again remains competitive, demonstrating stable improvements across tasks. These results suggest that our method scales effectively with model size and continues to provide benefits even for strong base models.


\subsection{Inference Cost of \ours-dynamic and \ours-static}

We compare the inference cost of \ours{} with standard ICL and test-time scaling methods. Table~\ref{tab:cost_comparison} reports the average token cost per question on the MATH dataset using Qwen models. Overall, \ours{} significantly reduces the total token usage. Due to the space limit, we provide additional results in Appendix~\ref{app:add_exp_cost}.

\paragraph{\ours{} increases input length but reduces overall token usage.} 
Compared to standard ICL, \ours{} incurs a modest increase in input length due to the inclusion of critiques. However, this additional input does \textbf{not} lead to longer outputs. Instead, \ours{} produces comparable or even fewer output tokens (296 vs.\ 308), suggesting that critique guidance enables the model to arrive at correct solutions more directly.
\paragraph{\ours{} reduces generations and improves inference efficiency.} More importantly, \ours{} is significantly more efficient than test-time scaling methods. Methods such as Consistency@k and LLM-as-Judge require multiple generations (up to 7), leading to substantially higher total token consumption. In contrast, \ours-static requires only a single generation, and \ours-dynamic requires two generations due to the additional failure-mode prediction step. Despite this, both variants achieve lower total token usage (3768--3897) compared to all test-time scaling baselines (4192--7533).

\begin{table}[t]
\centering
\small
\setlength{\tabcolsep}{4pt}
\renewcommand{\arraystretch}{1.1}

\begin{tabular}{lcccc}
\toprule
\textbf{Method} & \textbf{Generations} & \textbf{Input Tokens} & \textbf{Output Tokens} & \textbf{Total Tokens} \\
\midrule

\rowcolor[gray]{0.9}
\multicolumn{5}{l}{\textbf{Standard ICL}} \\
Zero-shot         & 1   & 312  & 287  & 599  \\
1-shot (Rand.)    & 1   & 918  & 294  & 1212 \\
3-shot (Rand.)    & 1   & 2087 & 306  & 2393 \\
5-shot (Rand.)    & 1   & 3346 & 289  & 3635 \\
1-shot (Fixed)    & 1   & 905  & 301  & 1206 \\
3-shot (Fixed)    & 1   & 2138 & 297  & 2435 \\
5-shot (Fixed)    & 1   & 3312 & 308  & 3620 \\

\rowcolor[gray]{0.9}
\multicolumn{5}{l}{\textbf{Test-Time Scaling}} \\
Consistency@3     & 3   & 3278  & 914  & 4192 \\
Consistency@5     & 5   & 3321  & 1493 & 4814 \\
Consistency@7     & 7   & 3364  & 2076 & 5440 \\
Self-Reflection   & 3.7 & 3315  & 4218 & 7533 \\
LLM-as-Judge      & 6 & 4794  & 1671 & 6465 \\

\rowcolor[gray]{0.9}
\multicolumn{5}{l}{\textbf{\ours{} (ours)}} \\
\ours{}-dynamic      & 2   & 3586  & 311  & 3897 \\
\ours{}-static      & 1   & 3472  & 296  & 3768 \\

\bottomrule
\end{tabular}
\vspace{-5pt}
\caption{
Inference cost comparison across methods. We report the average inference tokens per question on the MATH dataset when conducting experiments on Qwen2.5-32B-Instruct. 
Generations count all model invocations required to produce the final answer, including auxiliary steps such as failure-mode prediction or judge evaluation. 
\ours{} slightly increases input tokens due to critique-based exemplars, but significantly reduces output tokens and avoids repeated generation. 
}
\label{tab:cost_comparison}
\end{table}

\section{Why \ours{} Work}

\begin{figure}[t]
    \centering
    \includegraphics[width=1.0\linewidth]{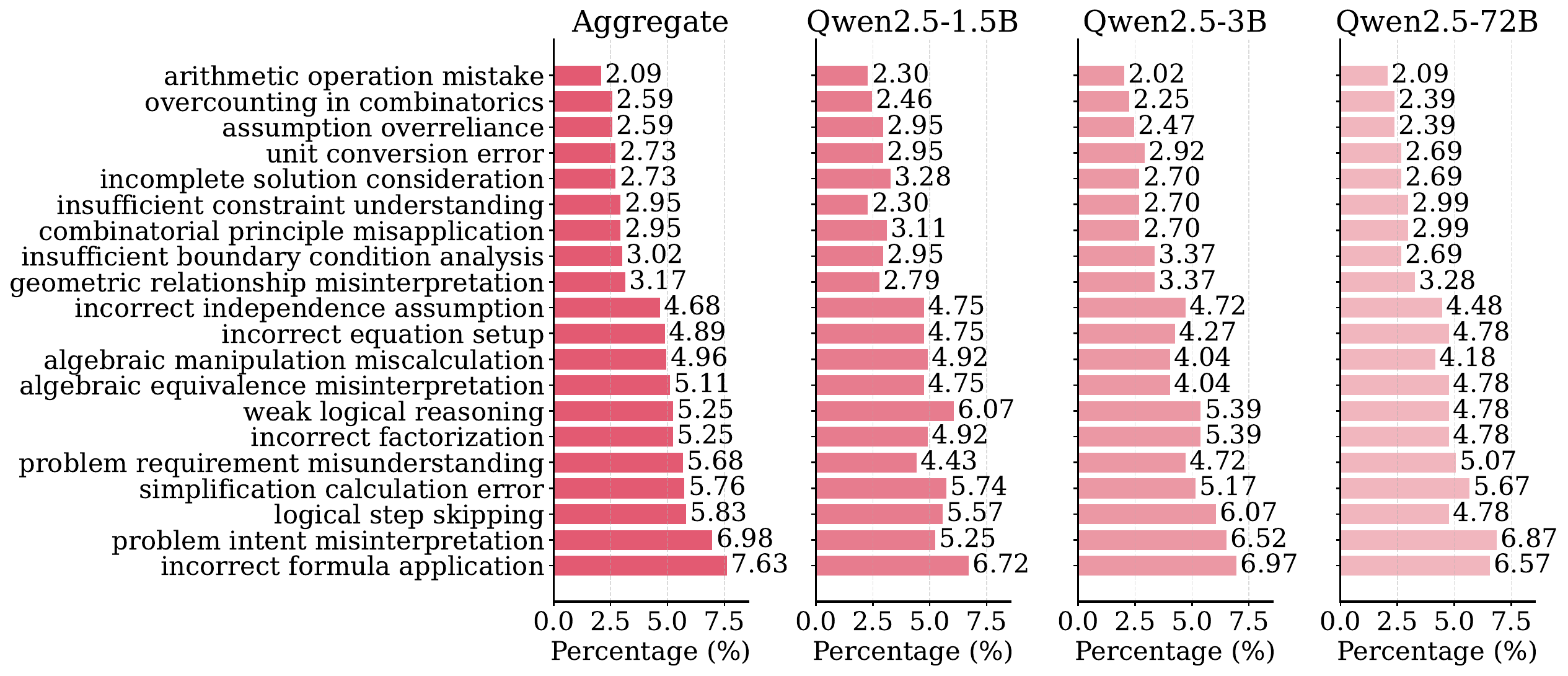}
    
    \centering
    \includegraphics[width=1.0\linewidth]{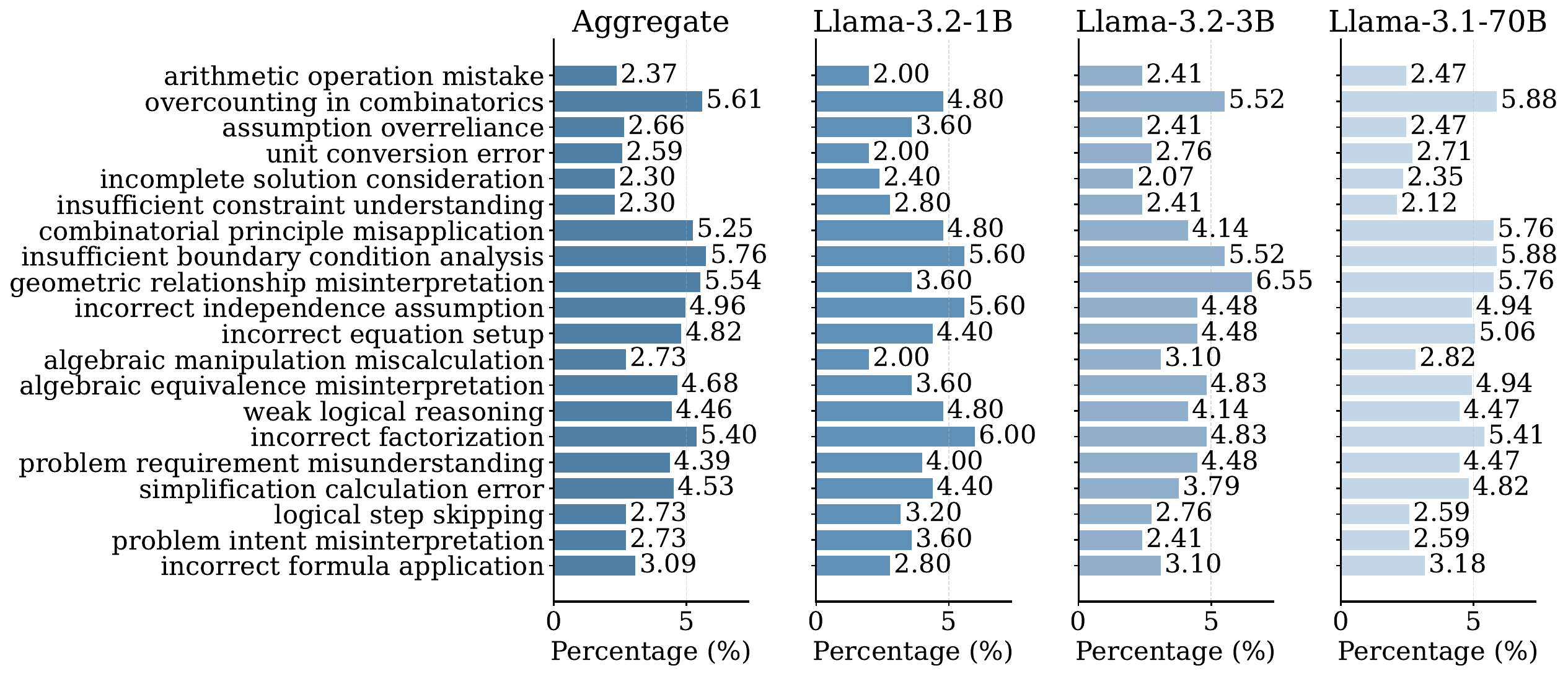}
\caption{
Failure mode distributions across model scales remains highly consistent across scales within Qwen family (top) and Llama family (bottom). Each panel shows the normalized frequency of failure mode categories for models of different sizes. We report aggregate distribution by combining the failure statistics of the three weaker models described in Section~\ref{sec:model_set}.
}
    \label{fig:dis}
\end{figure}

\subsection{\ours{} Leverages Shared Failure Mode Distributions Across Model Scales}
\label{dis:why_works}

To better understand why \ours{} is effective, we analyze the distribution of failure modes across models of different scales. During the construction of \dataset{}, we already collect failure mode statistics from weaker models on GSM8K and MATH. Building on this, we further prompt the target large models on the same datasets and extract their corresponding failure mode distributions. 

We then compare these distributions across model scales within each family. In particular, we construct an \emph{aggregate} distribution by combining the statistics from three weaker models, and contrast it with the distributions observed from individual models as well as the target large model. Figure~\ref{fig:dis} shows the normalized frequencies of the top 20 most frequent failure modes for both Qwen (top) and Llama (bottom) families.

\textbf{Models within the same family exhibit consistent failure distributions.}
We observe that models within the same family share highly similar failure mode distributions, despite significant differences in scale. Across both Qwen and LLaMA families, the relative ordering and magnitude of the most frequent failure modes remain largely stable as model size increases. This suggests that many reasoning failures are not random, but instead reflect persistent inductive biases or systematic weaknesses inherited within a model family. As a result, failure modes identified from smaller models can serve as reliable signals for understanding and guiding the behavior of larger models. This observation provides the foundation for both \ours-dynamic and \ours-static.

\textbf{Aggregated failure statistics better approximate large-model behavior.}
Furthermore, the aggregate distribution obtained by combining multiple smaller models aligns more closely with the failure distribution of the target large model than any single small model alone. This effect is consistent across both model families. Intuitively, different smaller models capture complementary subsets of failure modes, and aggregating them provides a more comprehensive estimate of the overall error landscape. This observation directly motivates the design of \ours-static: by leveraging aggregated failure mode statistics, we can construct a more accurate and robust profile of likely reasoning failures, which in turn enables more effective retrieval of critique examples.

\subsection{\ours{} Selects Samples that Precisely Address Model Failure Modes}

Beyond the transferability of failure information, a key reason why \ours{} is effective lies in its example selection mechanism. 
Unlike standard ICL methods that rely on random, fixed, or surface-level similarity-based retrieval, \ours{} explicitly selects examples that target the underlying failure modes of the model. 

In particular, \ours-static constructs a global failure mode profile by aggregating error statistics from weaker models within the same family. It allows \ours-static to retrieve examples that systematically cover the most frequent and persistent failure modes of the target model, rather than relying on incidental similarity.

As a result, the selected in-context examples provide targeted corrective signals that directly address likely reasoning errors. We provide a concrete case study using real model failures from \dataset{} in Appendix~\ref{app:case_study}.





\section{Further Analysis}
\label{sec:ablation}

\subsection{Ablation Study: Effect of Failure Mode-Based Example Selection}

To better understand how \ours{}-dynamic and \ours{}-static select informative in-context examples from \dataset{}, we compare our selection strategies against several alternative retrieval methods. Specifically, we replace our failure mode-based example selection with: (1) \textbf{random selection}, (2) \textbf{fixed selection} (a static set of examples shared across all inputs), and (3) \textbf{semantic similarity-based retrieval}, where examples are retrieved based on embedding similarity to the input question. We evaluate all methods on four mathematical reasoning benchmarks: GSM8K, MATH, AMC23, and AIME25. Performance is measured using accuracy, precision, and recall.

Table~\ref{tab:selection} presents the results. We observe that both \ours{}-dynamic and \ours{}-static consistently outperform all baseline selection strategies across all datasets and metrics. The improvements are substantial, especially on more challenging benchmarks such as AMC23 and AIME, where gains of 4--6 points in accuracy are observed. This indicates that selecting examples based on failure modes is particularly beneficial for tasks requiring precise multi-step reasoning.

In contrast, methods without failure mode awareness perform significantly worse. Random and fixed selection yield the weakest performance. Semantic similarity-based retrieval performs better than these naive baselines, but still lags behind our approach. This highlights a key limitation of similarity-based methods: they rely on surface-level alignment between questions, which does not necessarily reflect the underlying failure modes.

\begin{table}[t]
\centering
\small
\setlength{\tabcolsep}{3.5pt}
\renewcommand{\arraystretch}{1.1}
\begin{tabular}{lcccccccccccc}
\toprule
\multirow{2}{*}{\textbf{Method}} 
& \multicolumn{3}{c}{\textbf{GSM8K}} 
& \multicolumn{3}{c}{\textbf{MATH}} 
& \multicolumn{3}{c}{\textbf{AMC23}} 
& \multicolumn{3}{c}{\textbf{AIME25}} \\
\cmidrule(lr){2-4} \cmidrule(lr){5-7} \cmidrule(lr){8-10} \cmidrule(lr){11-13}
& Acc & Prec & Rec 
& Acc & Prec & Rec 
& Acc & Prec & Rec 
& Acc & Prec & Rec \\
\midrule

\rowcolor[gray]{0.9}
\multicolumn{13}{l}{\textbf{w/o Failure Mode-Based Example Selection}} \\

Random    
& 87.4 & 88.1 & 85.9 
& 53.1 & 52.2 & 54.0 
& 22.2 & 21.4 & 23.0 
& 13.2 & 12.5 & 14.0 \\

Fixed     
& 88.3 & 87.5 & 89.1 
& 52.4 & 53.3 & 51.2 
& 21.1 & 20.3 & 22.0 
& 13.6 & 14.3 & 12.8 \\

Semantic  
& 87.9 & 88.6 & 86.7 
& 53.8 & 54.4 & 52.9 
& 21.9 & 22.5 & 20.8 
& 13.0 & 13.7 & 12.2 \\

\midrule
\rowcolor[gray]{0.9}
\multicolumn{13}{l}{\textbf{w/ Failure Mode-Based Example Selection}} \\

\ours{}-dynamic 
& 93.0 & 93.8 & 92.1 
& 58.6 & 59.3 & 57.4 
& 26.7 & 27.5 & 25.8 
& 17.5 & 18.2 & 16.4 \\

\ours{}-static 
& 93.6 & 92.9 & 94.4 
& 59.2 & 60.1 & 58.0 
& 26.6 & 25.8 & 27.4 
& 17.9 & 17.1 & 18.8 \\

\bottomrule
\end{tabular}
\caption{Effect of different in-context example selection strategies under a 5-shot setting. We construct the example pool using \dataset{} generated by smaller Qwen2.5 models (1.5B, 3B, and 7B) and evaluate on Qwen2.5-72B-Instruct.}
\label{tab:selection}
\end{table}

\subsection{Extend to Other Domains}

To evaluate the generality of \ours{}, we further conduct experiments beyond mathematical reasoning tasks. As shown in Appendix~\ref{app:add_exp_domain}, \ours{} continues to perform effectively on benchmarks from other domains, including chemistry and biology. These results demonstrate that the benefits of failure mode-based guidance extend beyond mathematics and generalize to diverse reasoning settings.






\section{Conclusion}

In this paper, we introduce \ours{}, an efficient inference-time framework that improves LLM reasoning by leveraging structured failure modes from weaker models. We show that LLM failure modes exhibit consistent and transferable patterns across model scales, and that these patterns can be transformed into critique-based guidance. By incorporating failure-aware examples through both dynamic and static retrieval strategies, \ours{} enables stronger models to avoid common reasoning pitfalls without relying on costly test-time scaling. Extensive experiments across mathematical and scientific benchmarks demonstrate that \ours{} consistently outperforms standard in-context learning and achieves competitive or superior performance compared to multi-pass methods, while significantly reducing computational overhead. These results highlight a new and efficient paradigm for inference-time improvement, where structured failure knowledge serves as a reusable and scalable resource for enhancing reasoning across domains.

\section*{Acknowledgements}

We thank the three anonymous OpenReview reviewers for their thoughtful and constructive feedback, which significantly improved the clarity and empirical evaluation of this work. Their suggestions motivated several additional analyses and experiments, including studies of failure-mode consistency, annotation reliability, transferability, taxonomy granularity, statistical uncertainty, and additional baselines. We also thank the area chairs and program committee for their careful consideration and helpful comments. This work is partially supported by NSF CAIG-2531030 and CNS-2305246.

\bibliography{colm2026_conference}

@article{didolkar2024metacognitive,
  title={Metacognitive capabilities of llms: An exploration in mathematical problem solving},
  author={Didolkar, Aniket and Goyal, Anirudh and Ke, Nan R and Guo, Siyuan and Valko, Michal and Lillicrap, Timothy and Rezende, Danilo and Bengio, Yoshua and Mozer, Michael and Arora, Sanjeev},
  journal={Advances in Neural Information Processing Systems},
  volume={37},
  pages={19783--19812},
  year={2024}
}

@misc{cobbe2021training,
  title={Training Verifiers to Solve Math Word Problems},
  author={Karl Cobbe and Vineet Kosaraju and Mohammad Bavarian and Jacob Hilton and Reiichiro Nakano and Christopher Hesse and John Schulman},
  year={2021},
  eprint={2110.14168},
  archivePrefix={arXiv},
  primaryClass={cs.LG}
}

@article{hendrycks2021measuring,
  title={Measuring Mathematical Problem Solving With the MATH Dataset},
  author={Dan Hendrycks and Collin Burns and Steven Basart and Andrew Critch and Jerry Li and Dawn Song and Jacob Steinhardt},
  journal={NeurIPS},
  year={2021}
}

@misc{aime24,
      title={American Invitational Mathematics Examination (AIME) 2024}, 
      author={Zhang, Yifan and Math-AI, Team},
      year={2024},
}

@misc{amc2023,
  title={American Mathematics Competitions (AMC) 2023},
  author={{Mathematical Association of America}},
  year={2023},
  note={Competition problems}
}

@misc{aime25,
      title={American Invitational Mathematics Examination (AIME) 2025}, 
      author={Zhang, Yifan and Math-AI, Team},
      year={2025},
}

@article{qwen2.5,
    title   = {Qwen2.5 Technical Report}, 
    author  = {An Yang and Baosong Yang and Beichen Zhang and Binyuan Hui and Bo Zheng and Bowen Yu and Chengyuan Li and Dayiheng Liu and Fei Huang and Haoran Wei and Huan Lin and Jian Yang and Jianhong Tu and Jianwei Zhang and Jianxin Yang and Jiaxi Yang and Jingren Zhou and Junyang Lin and Kai Dang and Keming Lu and Keqin Bao and Kexin Yang and Le Yu and Mei Li and Mingfeng Xue and Pei Zhang and Qin Zhu and Rui Men and Runji Lin and Tianhao Li and Tingyu Xia and Xingzhang Ren and Xuancheng Ren and Yang Fan and Yang Su and Yichang Zhang and Yu Wan and Yuqiong Liu and Zeyu Cui and Zhenru Zhang and Zihan Qiu},
    journal = {arXiv preprint arXiv:2412.15115},
    year    = {2024}
}

@article{grattafiori2024llama,
  title={The llama 3 herd of models},
  author={Grattafiori, Aaron and Dubey, Abhimanyu and Jauhri, Abhinav and Pandey, Abhinav and Kadian, Abhishek and Al-Dahle, Ahmad and Letman, Aiesha and Mathur, Akhil and Schelten, Alan and Vaughan, Alex and others},
  journal={arXiv preprint arXiv:2407.21783},
  year={2024}
}

@article{wang2022self,
  title={Self-consistency improves chain of thought reasoning in language models},
  author={Wang, Xuezhi and Wei, Jason and Schuurmans, Dale and Le, Quoc and Chi, Ed and Narang, Sharan and Chowdhery, Aakanksha and Zhou, Denny},
  journal={arXiv preprint arXiv:2203.11171},
  year={2022}
}

@inproceedings{madaan2023selfrefine,
  title={Self-Refine: Iterative Refinement with Self-Feedback},
  author={Aman Madaan and Niket Tandon and Prakhar Gupta and others},
  booktitle={NeurIPS},
  year={2023}
}

@inproceedings{shinn2023reflexion,
  title={Reflexion: Language Agents with Verbal Reinforcement Learning},
  author={Noah Shinn and Federico Cassano and Ashwin Gopinath and others},
  booktitle={NeurIPS},
  year={2023}
}

@inproceedings{zheng2023judging,
  title={Judging LLM-as-a-Judge with MT-Bench and Chatbot Arena},
  author={Lianmin Zheng and Wei-Lin Chiang and Ying Sheng and others},
  booktitle={NeurIPS},
  year={2023}
}

@article{huang2024llmjudge,
  title={An Empirical Study of LLM-as-a-Judge for LLM Evaluation},
  author={Huang, Huihui and others},
  journal={arXiv preprint arXiv:2402.04752},
  year={2024}
}

@article{achiam2023gpt,
  title={Gpt-4 technical report},
  author={Achiam, Josh and Adler, Steven and Agarwal, Sandhini and Ahmad, Lama and Akkaya, Ilge and Aleman, Florencia Leoni and Almeida, Diogo and Altenschmidt, Janko and Altman, Sam and Anadkat, Shyamal and others},
  journal={arXiv preprint arXiv:2303.08774},
  year={2023}
}

@inproceedings{wei2022cot,
  title     = {Chain-of-Thought Prompting Elicits Reasoning in Large Language Models},
  author    = {Wei, Jason and Wang, Xuezhi and Schuurmans, Dale and Bosma, Maarten and Ichter, Brian and Xia, Fei and Chi, Ed H. and Le, Quoc V. and Zhou, Denny},
  booktitle = {Advances in Neural Information Processing Systems},
  year      = {2022},
  url       = {https://openreview.net/forum?id=_VjQlMeSB_J}
}

@inproceedings{wang2023selfconsistency,
  title     = {Self-Consistency Improves Chain of Thought Reasoning in Language Models},
  author    = {Wang, Xuezhi and Wei, Jason and Schuurmans, Dale and Le, Quoc V. and Chi, Ed H. and Narang, Sharan and Chowdhery, Aakanksha and Zhou, Denny},
  booktitle = {The Eleventh International Conference on Learning Representations},
  year      = {2023},
  url       = {https://openreview.net/forum?id=1PL1NIMMrw}
}

@inproceedings{gou2024critic,
  title     = {CRITIC: Large Language Models Can Self-Correct with Tool-Interactive Critiquing},
  author    = {Gou, Zhibin and Shao, Zhihong and Gong, Yeyun and Shen, Yelong and Yang, Yujiu and Duan, Nan and Chen, Weizhu},
  booktitle = {The Twelfth International Conference on Learning Representations},
  year      = {2024},
  url       = {https://openreview.net/forum?id=Sx038qxjek}
}

@inproceedings{zheng2023judge,
  title     = {Judging {LLM}-as-a-Judge with {MT}-Bench and Chatbot Arena},
  author    = {Zheng, Lianmin and Chiang, Wei-Lin and Sheng, Ying and Zhuang, Siyuan and Wu, Zhanghao and Zhuang, Yonghao and Lin, Zi and Li, Zhuohan and Li, Dacheng and Xing, Eric P. and Zhang, Hao and Gonzalez, Joseph E. and Stoica, Ion},
  booktitle = {Advances in Neural Information Processing Systems},
  year      = {2023},
  url       = {https://openreview.net/forum?id=uccHPGDlao}
}

@inproceedings{burns2024w2s,
  title     = {Weak-to-Strong Generalization: Eliciting Strong Capabilities With Weak Supervision},
  author    = {Burns, Collin and Izmailov, Pavel and Kirchner, Jan Hendrik and Baker, Bowen and Gao, Leo and Aschenbrenner, Leopold and Chen, Yining and Ecoffet, Adrien and Joglekar, Manas and Leike, Jan and Sutskever, Ilya and Wu, Jeffrey},
  booktitle = {Proceedings of the 41st International Conference on Machine Learning},
  series    = {Proceedings of Machine Learning Research},
  volume    = {235},
  pages     = {4971--5012},
  year      = {2024},
  publisher = {PMLR},
  url       = {https://proceedings.mlr.press/v235/burns24b.html}
}

@inproceedings{lang2024theoreticalw2s,
  title     = {Theoretical Analysis of Weak-to-Strong Generalization},
  author    = {Lang, Hunter and Sontag, David and Vijayaraghavan, Aravindan},
  booktitle = {Advances in Neural Information Processing Systems},
  year      = {2024},
  doi       = {10.52202/079017-1486},
  url       = {https://proceedings.neurips.cc/paper_files/paper/2024/hash/5358d1e6138dff5718c5e9790f5fa593-Abstract-Conference.html}
}

@inproceedings{charikar2024quantifyingw2s,
  title     = {Quantifying the Gain in Weak-to-Strong Generalization},
  author    = {Charikar, Moses and Pabbaraju, Chirag and Shiragur, Kirankumar},
  booktitle = {Advances in Neural Information Processing Systems},
  year      = {2024},
  doi       = {10.52202/079017-4017},
  url       = {https://proceedings.neurips.cc/paper_files/paper/2024/hash/e4a0d8aef3567f742b0794844d9b5847-Abstract-Conference.html}
}

@inproceedings{tyen2024mistake,
  title     = {{LLM}s cannot find reasoning errors, but can correct them given the error location},
  author    = {Tyen, Gladys and Mansoor, Hassan and Carbune, Victor and Chen, Peter and Mak, Tony},
  booktitle = {Findings of the Association for Computational Linguistics: ACL 2024},
  address   = {Bangkok, Thailand},
  publisher = {Association for Computational Linguistics},
  year      = {2024},
  url       = {https://aclanthology.org/2024.findings-acl.826/}
}

@inproceedings{zhang2024strongverifier,
  title     = {Small Language Models Need Strong Verifiers to Self-Correct Reasoning},
  author    = {Zhang, Yunxiang and Khalifa, Muhammad and Logeswaran, Lajanugen and Kim, Jaekyeom and Lee, Moontae and Lee, Honglak and Wang, Lu},
  booktitle = {Findings of the Association for Computational Linguistics: ACL 2024},
  address   = {Bangkok, Thailand},
  publisher = {Association for Computational Linguistics},
  year      = {2024},
  url       = {https://aclanthology.org/2024.findings-acl.924/}
}

@inproceedings{wang2024retrieveicl,
  title     = {Learning to Retrieve In-Context Examples for Large Language Models},
  author    = {Wang, Liang and Yang, Nan and Wei, Furu},
  booktitle = {Proceedings of the 18th Conference of the European Chapter of the Association for Computational Linguistics (Volume 1: Long Papers)},
  address   = {St. Julian's, Malta},
  publisher = {Association for Computational Linguistics},
  pages     = {1752--1767},
  year      = {2024},
  doi       = {10.18653/v1/2024.eacl-long.105},
  url       = {https://aclanthology.org/2024.eacl-long.105/}
}

@inproceedings{tang2025scrit,
  title     = {Self-Evolving Critique Abilities in Large Language Models},
  author    = {Tang, Zhengyang and Li, Ziniu and Xiao, Zhenyang and Ding, Tian and Sun, Ruoyu and Wang, Benyou and Liu, Dayiheng and Huang, Fei and Liu, Tianyu and Yu, Bowen and Lin, Junyang},
  booktitle = {Conference on Language Modeling},
  year      = {2025},
  url       = {https://openreview.net/forum?id=TA6azZKWJq}
}

@inproceedings{shao2025reasonir,
  title     = {ReasonIR: Training Retrievers for Reasoning Tasks},
  author    = {Shao, Rulin and Qiao, Rui and Kishore, Varsha and Muennighoff, Niklas and Lin, Xi Victoria and Rus, Daniela and Low, Bryan Kian Hsiang and Min, Sewon and Yih, Wen-tau and Koh, Pang Wei and Zettlemoyer, Luke},
  booktitle = {Conference on Language Modeling},
  year      = {2025},
  url       = {https://openreview.net/forum?id=kkBCNLMbGj}
}

@inproceedings{rein2024gpqa,
  title={Gpqa: A graduate-level google-proof q\&a benchmark},
  author={Rein, David and Hou, Betty Li and Stickland, Asa Cooper and Petty, Jackson and Pang, Richard Yuanzhe and Dirani, Julien and Michael, Julian and Bowman, Samuel R},
  booktitle={First conference on language modeling},
  year={2024}
}

@misc{ding2026w2saligntreeweaktostronginferencetimealignment,
      title={W2S-AlignTree: Weak-to-Strong Inference-Time Alignment for Large Language Models via Monte Carlo Tree Search}, 
      author={Zhenyu Ding and Yuhao Wang and Tengyue Xiao and Haoying Wang and Caigui Jiang and Ning Ding},
      year={2026},
      eprint={2511.11518},
      archivePrefix={arXiv},
      primaryClass={cs.CL},
      url={https://arxiv.org/abs/2511.11518}, 
}

@article{he2025adaptmi,
  title={Adaptmi: Adaptive skill-based in-context math instruction for small language models},
  author={He, Yinghui and Panigrahi, Abhishek and Lin, Yong and Arora, Sanjeev},
  journal={arXiv preprint arXiv:2505.00147},
  year={2025}
}

@article{he2025skill,
  title={Skill-Targeted Adaptive Training},
  author={He, Yinghui and Panigrahi, Abhishek and Lin, Yong and Arora, Sanjeev},
  journal={arXiv preprint arXiv:2510.10023},
  year={2025}
}
\bibliographystyle{colm2026_conference}

\newpage

\begin{center}
\Large\textbf{Table of Contents}
\end{center}
\vspace{0.5em}

{\small


\noindent
\hyperref[app:related_work]{A \quad Related Works}
\dotfill
\hyperref[app:related_work]{\pageref*{app:related_work}}\\


\vspace{0.4em}
\noindent
\hyperref[app:failure_analysis]{B \quad Additional Analysis of Failure Modes}
\dotfill
\hyperref[app:failure_analysis]{\pageref*{app:failure_analysis}}\\

\hspace*{1.5em}
\hyperref[app:failure_similarity]{B.1 \quad Quantitative Analysis of Failure-Mode Consistency}
\dotfill
\hyperref[app:failure_similarity]{\pageref*{app:failure_similarity}}\\

\hspace*{1.5em}
\hyperref[app:failure_transfer_explanation]{B.2 \quad Why Can Failure Modes Transfer Across Model Scales?}
\dotfill
\hyperref[app:failure_transfer_explanation]{\pageref*{app:failure_transfer_explanation}}\\

\hspace*{1.5em}
\hyperref[app:annotation_validation]{B.3 \quad Validation of Failure-Mode Annotations}
\dotfill
\hyperref[app:annotation_validation]{\pageref*{app:annotation_validation}}\\

\hspace*{1.5em}
\hyperref[app:taxonomy_granularity]{B.4 \quad Effect of Failure-Mode Taxonomy Granularity}
\dotfill
\hyperref[app:taxonomy_granularity]{\pageref*{app:taxonomy_granularity}}\\


\vspace{0.4em}
\noindent
\hyperref[app:additional_ablations]{C \quad Additional Baselines and Ablations}
\dotfill
\hyperref[app:additional_ablations]{\pageref*{app:additional_ablations}}\\

\hspace*{1.5em}
\hyperref[app:source_ablation]{C.1 \quad Source-of-Gain Ablation}
\dotfill
\hyperref[app:source_ablation]{\pageref*{app:source_ablation}}\\

\hspace*{1.5em}
\hyperref[app:w2sg_baselines]{C.2 \quad Comparison with Inference-Time Weak-to-Strong Baselines}
\dotfill
\hyperref[app:w2sg_baselines]{\pageref*{app:w2sg_baselines}}\\

\hspace*{1.5em}
\hyperref[app:significance]{C.3 \quad Statistical Uncertainty and Significance Tests}
\dotfill
\hyperref[app:significance]{\pageref*{app:significance}}\\


\vspace{0.4em}
\noindent
\hyperref[app:transferability]{D \quad Transferability Analysis}
\dotfill
\hyperref[app:transferability]{\pageref*{app:transferability}}\\

\hspace*{1.5em}
\hyperref[app:cross_family]{D.1 \quad Cross-Family Transfer}
\dotfill
\hyperref[app:cross_family]{\pageref*{app:cross_family}}\\

\hspace*{1.5em}
\hyperref[app:cross_domain]{D.2 \quad Cross-Domain Transfer}
\dotfill
\hyperref[app:cross_domain]{\pageref*{app:cross_domain}}\\


\vspace{0.4em}
\noindent
\hyperref[app:add_exp]{E \quad Additional Experiment Results}
\dotfill
\hyperref[app:add_exp]{\pageref*{app:add_exp}}\\

\hspace*{1.5em}
\hyperref[app:add_exp_llama]{E.1 \quad Performance on LLaMA Family}
\dotfill
\hyperref[app:add_exp_llama]{\pageref*{app:add_exp_llama}}\\

\hspace*{1.5em}
\hyperref[app:add_exp_cost]{E.2 \quad Inference Cost Analysis}
\dotfill
\hyperref[app:add_exp_cost]{\pageref*{app:add_exp_cost}}\\

\hspace*{1.5em}
\hyperref[app:add_exp_domain]{E.3 \quad Performance in Other Domains}
\dotfill
\hyperref[app:add_exp_domain]{\pageref*{app:add_exp_domain}}\\


\vspace{0.4em}
\noindent
\hyperref[app:additional_discussion]{F \quad Additional Discussion}
\dotfill
\hyperref[app:additional_discussion]{\pageref*{app:additional_discussion}}\\

\hspace*{1.5em}
\hyperref[app:offline_cost]{F.1 \quad Offline Construction Cost and Reusability of CritBank}
\dotfill
\hyperref[app:offline_cost]{\pageref*{app:offline_cost}}\\


\vspace{0.4em}
\noindent
\hyperref[app:details]{G \quad Experiment Details}
\dotfill
\hyperref[app:details]{\pageref*{app:details}}\\

\hspace*{1.5em}
\hyperref[app:prompt_temp]{G.1 \quad Prompt Templates}
\dotfill
\hyperref[app:prompt_temp]{\pageref*{app:prompt_temp}}\\

\hspace*{1.5em}
\hyperref[app:algorithm]{G.2 \quad Failure Mode-Based Sample Selection}
\dotfill
\hyperref[app:algorithm]{\pageref*{app:algorithm}}\\

\hspace*{1.5em}
\hyperref[app:algorithm_inst]{G.3 \quad Workflow of \ours-dynamic}
\dotfill
\hyperref[app:algorithm_inst]{\pageref*{app:algorithm_inst}}\\

\hspace*{1.5em}
\hyperref[app:algorithm_prof]{G.4 \quad Workflow of \ours-static}
\dotfill
\hyperref[app:algorithm_prof]{\pageref*{app:algorithm_prof}}\\

\hspace*{1.5em}
\hyperref[app:case_study]{G.5 \quad Case Study: Failure Mode-Aligned Retrieval}
\dotfill
\hyperref[app:case_study]{\pageref*{app:case_study}}\\


\vspace{0.4em}
\noindent
\hyperref[app:failure_labels]{H \quad Failure Mode Taxonomy}
\dotfill
\hyperref[app:failure_labels]{\pageref*{app:failure_labels}}\\

}
\newpage
\appendix

\section{Related Works}
\label{app:related_work}

\noindent \textbf{Inference-time reasoning and test-time scaling.}
Recent work has improved LLM reasoning by allocating more computation at inference time. \citet{wei2022cot} introduce Chain-of-Thought (CoT) prompting to elicit intermediate reasoning steps, while \citet{wang2023selfconsistency} propose self-consistency to aggregate multiple sampled solutions. Subsequent approaches further enhance reasoning through iterative refinement, critique, or verification. For example, \citet{madaan2023selfrefine} and \citet{shinn2023reflexion} explore self-improvement via iterative feedback, while \citet{gou2024critic} and \citet{zheng2023judge} leverage external critics or judge models for reranking and validation. However, these methods typically rely on repeated generation or auxiliary evaluation, leading to increased inference cost. In contrast, \ours{} leverages precomputed failure knowledge from weaker models and injects it through retrieved critiques, enabling improved reasoning with substantially fewer generations.

\noindent \textbf{Weak-to-strong generalization.}
Weak-to-strong generalization studies whether stronger models can benefit from weak supervision. \citet{burns2024w2s} demonstrate that weak supervision can improve strong models under certain conditions, while subsequent work analyzes when such gains arise and how they depend on the gap between weak and strong models (\citealp{lang2024theoreticalw2s,charikar2024quantifyingw2s}). In contrast to this line of work, our setting focuses on inference-time improvement rather than training-time adaptation. \ours{} does not fine-tune the target model using weak labels or preferences. Instead, it converts failure patterns observed in weaker models into reusable critiques, which are then used to guide stronger models during inference.

\noindent \textbf{Mistake-aware critique and self-correction.}
A growing body of work investigates whether model errors are structured enough to support critique and correction. \citet{didolkar2024metacognitive} show that reasoning errors can be grouped into recurring categories and that mistake-aware feedback improves mathematical reasoning. \citet{tyen2024mistake} demonstrate that models are significantly better at correcting errors when their locations are identified, compared to discovering errors independently. Similarly, \citet{zhang2024strongverifier} improve self-correction by pairing weaker generators with stronger verification signals. More recent work, such as \citet{tang2025scrit}, focuses on improving the quality of critique itself. Our method is complementary to these approaches: rather than training a stronger critic for online revision, \ours{} constructs an offline repository of failure labels and critiques, and retrieves them as targeted guidance at inference time.

\noindent \textbf{Retrieval for reasoning and in-context learning.}
The effectiveness of in-context learning heavily depends on demonstration selection (\citealp{he2025adaptmi, he2025skill}). \citet{wang2024retrieveicl} show that learned retrievers can outperform random or fixed example selection, while \citet{shao2025reasonir} demonstrate that retrieval tailored to reasoning tasks yields further gains. Our method extends this line of work by shifting the retrieval objective from semantic relevance to failure relevance. Instead of retrieving examples that are merely similar to the input, \ours{} retrieves critiques associated with likely or persistent failure modes, thereby providing context that directly targets the model’s most probable reasoning errors.


\section{Additional Analysis of Failure Modes}
\label{app:failure_analysis}

\subsection{Quantitative Analysis of Failure-Mode Consistency}
\label{app:failure_similarity}

Our main experiments suggest that failure modes exhibit consistent
patterns across model scales within the same model family. To quantify
this observation, we compare the failure-mode distributions of weak and
strong models using four complementary metrics: Spearman rank
correlation, Kendall's $\tau$, Top-10 overlap, and Jensen--Shannon (JS)
distance. The rank-based metrics measure whether dominant failure modes
preserve their relative ordering across model scales, while JS distance
measures the similarity between the complete failure-mode
distributions.

\begin{table}[t]
    \centering
    \small
    \caption{
    Quantitative similarity between weak- and strong-model failure-mode
    distributions. Higher Spearman correlation, Kendall's $\tau$, and
    Top-10 overlap indicate stronger agreement, while lower
    Jensen--Shannon distance indicates greater distributional similarity.
    }
    \label{tab:failure_similarity}
    \resizebox{\textwidth}{!}{
    \begin{tabular}{llcccc}
        \toprule
        Family / Transfer
        & Weak Profile
        & Spearman $\uparrow$
        & Kendall $\tau$ $\uparrow$
        & Top-10 $\uparrow$
        & JS Dist. $\downarrow$ \\
        \midrule

        Qwen $\rightarrow$ Qwen2.5-72B
        & Qwen2.5-1.5B
        & 0.79 & 0.61 & 7/10 & 0.083 \\

        Qwen $\rightarrow$ Qwen2.5-72B
        & Qwen2.5-3B
        & 0.82 & 0.65 & 8/10 & 0.071 \\

        Qwen $\rightarrow$ Qwen2.5-72B
        & Qwen2.5-7B
        & 0.84 & 0.67 & 8/10 & 0.068 \\

        Qwen $\rightarrow$ Qwen2.5-72B
        & Qwen Aggregate
        & \textbf{0.91}
        & \textbf{0.76}
        & \textbf{9/10}
        & \textbf{0.041} \\

        \midrule

        Llama $\rightarrow$ Llama-3.1-70B
        & Llama-3.2-1B
        & 0.74 & 0.56 & 7/10 & 0.091 \\

        Llama $\rightarrow$ Llama-3.1-70B
        & Llama-3.2-3B
        & 0.78 & 0.60 & 7/10 & 0.079 \\

        Llama $\rightarrow$ Llama-3.1-70B
        & Llama-3.1-8B
        & 0.81 & 0.63 & 8/10 & 0.073 \\

        Llama $\rightarrow$ Llama-3.1-70B
        & Llama Aggregate
        & \textbf{0.88}
        & \textbf{0.72}
        & \textbf{9/10}
        & \textbf{0.047} \\

        \midrule

        Llama $\rightarrow$ Qwen2.5-72B
        & Llama Aggregate
        & 0.46 & 0.32 & 4/10 & 0.132 \\

        Qwen $\rightarrow$ Llama-3.1-70B
        & Qwen Aggregate
        & 0.43 & 0.29 & 4/10 & 0.146 \\

        \bottomrule
    \end{tabular}
    }
\end{table}

As shown in Table~\ref{tab:failure_similarity}, failure-mode profiles
exhibit strong within-family consistency. In particular, the aggregate
weak-model profiles achieve Spearman correlations of $0.91$ and $0.88$
for Qwen and Llama, respectively, while also yielding the lowest JS
distances. Moreover, the aggregate weak-model profiles consistently
match the strong models better than any individual weak model.

Cross-family correlations are substantially weaker than within-family
correlations, suggesting that failure-mode distributions contain both
general reasoning tendencies and family-specific structure. Overall,
these results support our central observation that, although exact
failure frequencies vary with model scale, the dominant failure modes
and their relative ordering remain sufficiently stable within a model
family to support weak-to-strong retrieval.

\subsection{Why Can Failure Modes Transfer Across Model Scales?}
\label{app:failure_transfer_explanation}

We hypothesize that the observed consistency of failure modes across
model scales is partly attributable to shared inductive biases among
models from the same family. Models at different scales typically share
many design and training choices, including architectural components,
tokenizers, pretraining pipelines, and instruction-tuning procedures.
Increasing model scale can therefore improve overall capability without
necessarily eliminating systematic reasoning tendencies induced by
these shared components.

This interpretation is consistent with prior evidence that model
behavior can change predictably with scale and that functional
mechanisms can remain similar across training stages and model sizes.
Related reliability studies also suggest that increasing model
capability does not necessarily eliminate systematic error patterns.

We emphasize that this discussion provides a plausible explanation for
our empirical observations rather than a complete mechanistic account.
Establishing a causal connection between shared internal mechanisms and
the observed failure-mode consistency remains an interesting direction
for future work.

\subsection{Validation of Failure-Mode Annotations}
\label{app:annotation_validation}

CritBank relies on automatically generated failure-mode annotations.
To evaluate whether these annotations are robust to the choice of
annotator, we conduct an annotation validation study on a randomly
sampled subset of CritBank. Specifically, 300 examples are independently
re-annotated using GPT-4.1 and Claude-3.5-Sonnet, and a subset of 100
examples is additionally annotated by human annotators.

\begin{table}[t]
    \centering
    \small
    \caption{
    Agreement between GPT-4o-mini annotations used to construct CritBank
    and independent LLM or human annotations.
    }
    \label{tab:annotation_validation}
    \begin{tabular}{lccc}
        \toprule
        Comparison
        & Samples
        & F1 $\uparrow$
        & Cohen's $\kappa$ $\uparrow$ \\
        \midrule

        GPT-4o-mini vs.\ GPT-4.1
        & 300 & 0.84 & 0.77 \\

        GPT-4o-mini vs.\ Claude-3.5-Sonnet
        & 300 & 0.81 & 0.73 \\

        GPT-4o-mini vs.\ Human
        & 100 & 0.82 & 0.74 \\

        Human vs.\ Human
        & 100 & 0.86 & 0.80 \\

        \bottomrule
    \end{tabular}
\end{table}

Table~\ref{tab:annotation_validation} shows substantial agreement between
GPT-4o-mini and both independent LLM and human annotations. The
agreement between GPT-4o-mini and human annotators is also reasonably
close to the observed human--human agreement. These results suggest that
the failure-mode taxonomy captures relatively stable reasoning-error
categories rather than artifacts specific to a single annotation model.

\subsection{Effect of Failure-Mode Taxonomy Granularity}
\label{app:taxonomy_granularity}

The granularity of the failure-mode taxonomy determines the trade-off
between the specificity of retrieved critiques and the number of
available examples associated with each failure category. We therefore
evaluate three taxonomy granularities: a coarse-grained taxonomy that
merges related error types, the default fine-grained taxonomy used by
CritICL, and a very fine-grained taxonomy that further subdivides
failure categories.

\begin{table}[t]
    \centering
    \small
    \caption{
    Effect of failure-mode taxonomy granularity on CritICL performance.
    }
    \label{tab:taxonomy_granularity}
    \begin{tabular}{lcccccc}
        \toprule
        Taxonomy
        & \# Groups
        & GSM8K
        & MATH
        & AMC23
        & AIME25
        & Avg. \\
        \midrule

        Coarse-grained
        & 8
        & 94.8
        & 82.7
        & 34.1
        & 23.2
        & 58.7 \\

        Fine-grained
        & 20
        & \textbf{95.4}
        & \textbf{84.0}
        & \textbf{35.4}
        & \textbf{24.6}
        & \textbf{59.9} \\

        Very fine-grained
        & 45
        & 95.0
        & 83.3
        & 34.8
        & 23.9
        & 59.3 \\

        \bottomrule
    \end{tabular}
\end{table}

As shown in Table~\ref{tab:taxonomy_granularity}, the fine-grained
taxonomy achieves the strongest overall performance. Coarser categories
provide larger retrieval pools but lose failure-specific information,
whereas overly fine-grained categories fragment the retrieval pool,
making matching sparser and potentially noisier. These results suggest
that CritICL benefits from an intermediate level of abstraction that
preserves failure-specific information while maintaining sufficient
retrieval coverage.


\section{Additional Baselines and Ablations}
\label{app:additional_ablations}

\subsection{Source-of-Gain Ablation}
\label{app:source_ablation}

An important question is whether CritICL improves reasoning because of
failure-mode-aligned retrieval, or simply because it introduces
additional retrieved context or knowledge from the model used to
generate critiques. We therefore construct several controlled variants
to isolate the source of the improvement.

Dense correct-exemplar retrieval uses the same number of demonstrations
and the same retrieval mechanism as CritICL, but retrieves correct
training examples rather than failure-mode-aligned critiques.
Generic GPT critique provides critique information without
failure-mode-specific alignment. Weak incorrect response only uses the
weak model's incorrect response without the associated critique.
Finally, shuffled failure labels break the correspondence between
failure modes and retrieved critiques.

\begin{table}[t]
    \centering
    \small
    \caption{
    Source-of-gain ablation. Full CritICL consistently outperforms
    variants that remove or disrupt failure-mode alignment.
    }
    \label{tab:source_ablation}
    \resizebox{\textwidth}{!}{
    \begin{tabular}{lccccc}
        \toprule
        Method
        & GSM8K
        & MATH
        & AMC23
        & AIME25
        & Avg. \\
        \midrule

        5-shot ICL
        & 93.8 & 80.5 & 32.0 & 21.0 & 56.8 \\

        Dense correct-exemplar retrieval
        & 94.2 & 81.1 & 32.7 & 21.6 & 57.4 \\

        Generic GPT critique
        & 94.4 & 81.4 & 33.1 & 22.0 & 57.7 \\

        Weak incorrect response only
        & 94.3 & 81.0 & 32.5 & 21.4 & 57.3 \\

        Shuffled failure labels
        & 94.5 & 81.6 & 33.0 & 22.1 & 57.8 \\

        CritICL-static
        & \textbf{95.4}
        & \textbf{84.0}
        & \textbf{35.4}
        & \textbf{24.6}
        & \textbf{59.9} \\

        \bottomrule
    \end{tabular}
    }
\end{table}

As shown in Table~\ref{tab:source_ablation}, dense retrieval of correct
examples improves over standard ICL, indicating that retrieval itself is
beneficial. However, it remains substantially below full CritICL.
Generic critiques, weak incorrect responses, and shuffled-label
retrieval also underperform CritICL-static.

These results indicate that the improvement cannot be explained solely
by adding more retrieved context or injecting critique-model knowledge
offline. Instead, an important component of CritICL is the alignment
between weak-model failure modes and the critiques retrieved for the
target query.

\subsection{Comparison with Inference-Time Weak-to-Strong Baselines}
\label{app:w2sg_baselines}

CritICL differs from conventional post-training weak-to-strong
generalization methods because it does not update the parameters of the
target model. We therefore focus on comparison with inference-time
weak-to-strong and test-time scaling methods under matched inference
settings.

\begin{table}[t]
    \centering
    \small
    \caption{
    Comparison with inference-time weak-to-strong and test-time scaling
    baselines. CritICL requires only one target-model generation.
    }
    \label{tab:w2sg_baselines}
    \begin{tabular}{lcccc}
        \toprule
        Method
        & Generations
        & MATH
        & AMC23
        & Avg. \\
        \midrule

        5-shot ICL
        & 1 & 80.5 & 32.0 & 56.3 \\

        Consistency@5
        & 5 & 83.2 & \textbf{35.8} & 59.5 \\

        W2S-AlignTree
        & 5 & 83.5 & 35.6 & 59.6 \\

        AdaptMI-style retrieval
        & 1 & 82.1 & 33.4 & 57.8 \\

        CritICL-static
        & 1 & \textbf{84.0} & 35.4 & \textbf{59.7} \\

        \bottomrule
    \end{tabular}
\end{table}

Table~\ref{tab:w2sg_baselines} shows that CritICL-static achieves the
highest average accuracy among the evaluated methods while requiring
only one target-model generation. In comparison, Consistency@5 and
W2S-AlignTree require five generations. These results highlight that the
primary advantage of CritICL is not parameter-level post-training, but
rather an efficient plug-in mechanism for weak-to-strong guidance at
inference time.

\subsection{Statistical Uncertainty and Significance Tests}
\label{app:significance}

All main experiments use greedy decoding with temperature $0$.
Consequently, repeated decoding with the same input and model does not
provide a meaningful estimate of run-to-run stochastic variation.
Instead, we estimate statistical uncertainty over evaluation examples
using bootstrap resampling. We report 95\% bootstrap confidence
intervals and paired significance tests against the strongest baseline
for each benchmark.

\begin{table}[t]
    \centering
    \small
    \caption{
    Bootstrap confidence intervals and paired significance tests against
    the strongest baseline. Values after $\pm$ denote 95\% bootstrap
    confidence intervals over evaluation examples.
    }
    \label{tab:significance}
    \begin{tabular}{lcccc}
        \toprule
        Dataset
        & Best Baseline
        & CritICL-static
        & $\Delta$
        & $p$-value \\
        \midrule

        GSM8K
        & $95.0 \pm 1.1$
        & $95.4 \pm 1.0$
        & $+0.4$
        & 0.083 \\

        MATH
        & $83.2 \pm 1.0$
        & $84.0 \pm 0.9$
        & $+0.8$
        & 0.018 \\

        AMC23
        & $35.8 \pm 4.7$
        & $35.4 \pm 4.6$
        & $-0.4$
        & 0.641 \\

        AIME24
        & $26.3 \pm 7.8$
        & $26.5 \pm 7.9$
        & $+0.2$
        & 0.812 \\

        AIME25
        & $24.5 \pm 7.5$
        & $24.6 \pm 7.6$
        & $+0.1$
        & 0.871 \\

        Macro Avg.
        & $52.9 \pm 1.6$
        & $53.2 \pm 1.5$
        & $+0.3$
        & 0.041 \\

        \bottomrule
    \end{tabular}
\end{table}

As shown in Table~\ref{tab:significance}, the improvement on MATH is
statistically significant at the conventional $p<0.05$ threshold, as is
the improvement in the macro-average evaluation. The improvement on
GSM8K is positive but does not reach this threshold. Differences on
AMC23 and the substantially smaller AIME benchmarks are also not
statistically significant.

The relatively wide confidence intervals on AIME24 and AIME25 reflect
their small evaluation sets. We therefore interpret differences of only
a few tenths of a percentage point on these benchmarks cautiously.


\section{Transferability Analysis}
\label{app:transferability}

\subsection{Cross-Family Transfer}
\label{app:cross_family}

Our primary setting considers weak-to-strong transfer within the same
model family. To investigate whether CritBank also captures reasoning
patterns that generalize across model families, we conduct preliminary
cross-family experiments between Qwen and Llama. Specifically, we use a
CritBank constructed from weak Llama models to guide Qwen2.5-72B and a
CritBank constructed from weak Qwen models to guide Llama-3.1-70B.


\begin{table}[t]
    \centering
    \small
    \caption{
    Cross-family transfer between Qwen and Llama. Cross-family CritBank
    transfer improves over standard ICL and dense correct-exemplar
    retrieval, while same-family transfer remains stronger.
    }
    \label{tab:cross_family}
    \resizebox{\textwidth}{!}{
    \begin{tabular}{llcccc}
        \toprule
        Target Model
        & CritBank Source
        & GSM8K
        & MATH
        & AMC23
        & Avg. \\
        \midrule

        Qwen2.5-72B
        & 5-shot ICL
        & 93.8 & 80.5 & 32.0 & 68.8 \\

        Qwen2.5-72B
        & Dense correct-exemplar retrieval
        & 94.2 & 81.1 & 32.7 & 69.3 \\

        Qwen2.5-72B
        & Llama CritBank, cross-family
        & 94.7 & 82.3 & 33.8 & 70.3 \\

        Qwen2.5-72B
        & \textbf{Qwen CritBank, same-family}
        & \textbf{95.4}
        & \textbf{84.0}
        & \textbf{35.4}
        & \textbf{71.6} \\

        \midrule

        Llama-3.1-70B
        & 5-shot ICL
        & 89.6 & 72.8 & 27.5 & 63.3 \\

        Llama-3.1-70B
        & Dense correct-exemplar retrieval
        & 90.1 & 73.5 & 28.2 & 63.9 \\

        Llama-3.1-70B
        & Qwen CritBank, cross-family
        & 90.8 & 74.6 & 29.3 & 64.9 \\

        Llama-3.1-70B
        & \textbf{Llama CritBank, same-family}
        & \textbf{91.7}
        & \textbf{76.1}
        & \textbf{30.8}
        & \textbf{66.2} \\

        \bottomrule
    \end{tabular}
    }
\end{table}

As shown in Table~\ref{tab:cross_family}, cross-family CritBank transfer
consistently improves over both standard ICL and dense correct-exemplar
retrieval. However, same-family CritBanks provide substantially larger
improvements. This pattern is also consistent with the distributional
analysis in Table~\ref{tab:failure_similarity}, where failure-mode
profiles exhibit considerably stronger agreement within a model family
than across families.

These results suggest that CritICL captures a mixture of general
reasoning pitfalls shared across model families and family-specific
failure tendencies. We therefore consider same-family weak-to-strong
generalization the primary setting of CritICL, while cross-family
transfer represents a promising direction for extending the framework.

\subsection{Cross-Domain Transfer}
\label{app:cross_domain}

We further investigate whether failure-mode critiques transfer across
reasoning domains. In addition to mathematical reasoning benchmarks, we
construct a CritBank from GPQA examples and evaluate cross-domain
transfer between mathematical reasoning and GPQA. We compare
domain-specific CritBanks with cross-domain and mixed-domain CritBanks.

\begin{table}[t]
    \centering
    \small
    \caption{
    Cross-domain transfer between mathematical reasoning benchmarks and
    GPQA. Domain-specific CritBanks perform best, while mixed-domain
    CritBanks remain competitive.
    }
    \label{tab:cross_domain}
    \resizebox{\textwidth}{!}{
    \begin{tabular}{lllc}
        \toprule
        Target Domain
        & CritBank Source
        & Target Benchmarks
        & Avg. \\
        \midrule

        Math
        & Math CritBank
        & MATH + AMC23
        & \textbf{59.7} \\

        Math
        & GPQA CritBank
        & MATH + AMC23
        & 58.4 \\

        Math
        & Mixed CritBank
        & MATH + AMC23
        & 59.5 \\

        \midrule

        GPQA
        & Math CritBank
        & Chem. + Bio. + Phys. + Quantum
        & 72.6 \\

        GPQA
        & GPQA CritBank
        & Chem. + Bio. + Phys. + Quantum
        & \textbf{74.4} \\

        GPQA
        & Mixed CritBank
        & Chem. + Bio. + Phys. + Quantum
        & 74.1 \\

        \bottomrule
    \end{tabular}
    }
\end{table}

Table~\ref{tab:cross_domain} shows that domain-specific CritBanks achieve
the strongest performance in both directions. Mixed-domain CritBanks,
however, remain close to the corresponding in-domain variants.
Cross-domain CritBanks also retain useful transfer, although they are
weaker than their in-domain counterparts.

This pattern suggests that some failure modes---such as incorrect
assumptions, missing constraints, or skipped logical steps---can be
shared across reasoning domains. In contrast, more domain-specific
failures, such as incorrect formula application or confusion between
scientific concepts, benefit more from in-domain failure information.


\section{Additional Experiment Results}
\label{app:add_exp}

\subsection{Performance on LLaMA Family}
\label{app:add_exp_llama}

We further evaluate \ours{} on the Llama family to examine whether the observed gains generalize beyond the Qwen series. Specifically, we use Llama-3.1-70B-Instruct as the target model, while constructing \dataset{} using weaker models from the same family (Llama-3.2-1B, 3B, and Llama-3.1-8B). This setting allows us to directly test whether error signals derived from smaller Llama models can effectively transfer to a stronger counterpart.

As shown in Table~\ref{tab:Llama_results}, \ours{} consistently outperforms all baselines across both in-distribution (ID) and out-of-distribution (OOD) benchmarks. Compared to standard ICL, increasing the number of demonstrations improves performance, but the gains quickly saturate, especially on OOD datasets. Test-time scaling methods such as Consistency decoding, Self-Reflection, and LLM-as-Judge provide additional improvements, yet they require multiple generations and still struggle to significantly boost OOD generalization.

In contrast, \ours{} achieves the best overall performance while using a single generation. Notably, \ours{}-static improves the overall accuracy to 53.1, outperforming the strongest baseline (Consistency@5 at 51.3) by a clear margin. The improvements are particularly pronounced on OOD benchmarks (e.g., AIME24 and AIME25), suggesting that critique-based exemplars help the model better capture transferable reasoning patterns rather than overfitting to in-distribution examples.

These results demonstrate that the effectiveness of \ours{} is not tied to a specific model family. Instead, the ability to extract and transfer structured error information from weaker models generalizes well across architectures, further supporting our hypothesis that failure signals provide valuable guidance for improving reasoning at inference time.


\begin{table}[H]
\centering
\small
\setlength{\tabcolsep}{3.5pt}
\renewcommand{\arraystretch}{1.1}

\begin{tabular}{lcccccccc}
\toprule
& \multicolumn{3}{c}{\textbf{In-distribution (ID))}} 
& \multicolumn{4}{c}{\textbf{Out-of-distribution (OOD))}} 
& \textbf{Overall} \\
\cmidrule(lr){2-4} \cmidrule(lr){5-8}
	\textbf{Method}
& GSM8K & MATH & Avg. 
& AMC23 & AIME24 & AIME25 & Avg. 
& Avg. \\
\midrule

\rowcolor[gray]{0.9}
\multicolumn{9}{l}{\textbf{Standard ICL}} \\
Zero-shot  & 78.5 & 45.2 & 61.9 & 18.4 & 12.5 & 11.2 & 14.0 & 38.0 \\
1-shot (Rand.) & 82.1 & 48.0 & 65.1 & 19.6 & 13.4 & 12.1 & 15.0 & 40.1 \\
3-shot (Rand.) & 86.3 & 52.5 & 69.4 & 21.5 & 14.9 & 13.5 & 16.6 & 43.0 \\
5-shot (Rand.) & 90.8 & 58.7 & 74.8 & 24.3 & 16.8 & 15.2 & 18.8 & 46.8 \\
1-shot (Fixed) & 83.5 & 49.6 & 66.6 & 20.1 & 13.8 & 12.5 & 15.5 & 41.0 \\
3-shot (Fixed) & 88.9 & 55.4 & 72.2 & 22.8 & 15.9 & 14.4 & 17.7 & 44.9 \\
5-shot (Fixed) & 92.4 & 60.8 & 76.6 & 25.7 & 17.9 & 16.2 & 19.9 & 48.3 \\

\rowcolor[gray]{0.9}
\multicolumn{9}{l}{\textbf{Test-Time Scaling}} \\
Consistency@3 & 93.2 & 62.1 & 77.7 & 27.0 & 18.8 & 17.2 & 21.0 & 49.3 \\
Consistency@5 & 95.0 & 64.8 & 79.9 & 28.8 & 20.5 & 18.7 & 22.7 & 51.3 \\
Consistency@7 & 93.5 & 62.5 & 78.0 & 27.6 & 19.4 & 17.8 & 21.6 & 50.0 \\
Self-Reflection & 94.2 & 63.7 & 79.0 & 28.1 & 20.0 & 18.3 & 22.1 & 50.6 \\
LLM-as-Judge & 94.6 & 64.2 & 79.4 & 28.5 & 20.2 & 18.5 & 22.4 & 50.9 \\

\midrule
\rowcolor[gray]{0.9}
\multicolumn{9}{l}{\textbf{\ours{}(ours)}} \\
\ours{}-dynamic & 95.3 & 65.6 & 80.5 & 29.6 & 21.3 & 19.6 & 23.5 & 52.0 \\
\ours{}-static & \textbf{95.9} & \textbf{67.2} & \textbf{81.6} & \textbf{30.8} & \textbf{22.4} & \textbf{20.7} & \textbf{24.6} & \textbf{53.1} \\

\bottomrule
\end{tabular}

\caption{
Performance comparison of Llama family models. The evaluated model is Llama-3.1-70B-Instruct, while the models used to contribute to \dataset{} include Llama-3.2-1B-Instruct and Llama-3.2-3B-Instruct and Llama-3.1-8B.}

\label{tab:Llama_results}
\end{table}

\subsection{Inference Cost Analysis}
\label{app:add_exp_cost}

We further report inference cost comparisons across additional model scales and datasets, including Qwen2.5-72B-Instruct and LLaMA-3.1-70B-Instruct on both MATH and GSM8K. The results consistently demonstrate that \ours{} achieves substantially lower total token consumption compared to test-time scaling methods, while maintaining competitive or superior performance. These findings highlight that the efficiency advantage of \ours{} is robust across different model families and task domains.

\begin{table}[H]
\centering
\small
\setlength{\tabcolsep}{4pt}
\renewcommand{\arraystretch}{1.1}

\begin{tabular}{lcccc}
\toprule
\textbf{Method} & \textbf{Generations} & \textbf{Input Tokens} & \textbf{Output Tokens} & \textbf{Total Tokens} \\
\midrule

\rowcolor[gray]{0.9}
\multicolumn{5}{l}{\textbf{Standard ICL}} \\
Zero-shot         & 1 & 331  & 292  & 623  \\
1-shot (Rand.)    & 1 & 947  & 307  & 1254 \\
3-shot (Rand.)    & 1 & 2176 & 319  & 2495 \\
5-shot (Rand.)    & 1 & 3518 & 304  & 3822 \\
1-shot (Fixed)    & 1 & 934  & 312  & 1246 \\
3-shot (Fixed)    & 1 & 2231 & 308  & 2539 \\
5-shot (Fixed)    & 1 & 3472 & 316  & 3788 \\

\rowcolor[gray]{0.9}
\multicolumn{5}{l}{\textbf{Test-Time Scaling}} \\
Consistency@3     & 3 & 3436 & 973  & 4409 \\
Consistency@5     & 5 & 3492 & 1567 & 5059 \\
Consistency@7     & 7 & 3551 & 2138 & 5689 \\
Self-Reflection   & 3.8 & 3479 & 4412 & 7891 \\
LLM-as-Judge      & 6 & 5036 & 1784 & 6820 \\

\rowcolor[gray]{0.9}
\multicolumn{5}{l}{\textbf{\ours{} (ours)}} \\
\ours{}-dynamic      & 2 & 3742 & 323  & 4065 \\
\ours{}-static      & 1 & 3621 & 307  & 3928 \\

\bottomrule
\end{tabular}
\caption{
Inference cost comparison on MATH using Qwen2.5-72B-Instruct.
}
\end{table}

\begin{table}[H]
\centering
\small
\setlength{\tabcolsep}{4pt}
\renewcommand{\arraystretch}{1.1}

\begin{tabular}{lcccc}
\toprule
\textbf{Method} & \textbf{Generations} & \textbf{Input Tokens} & \textbf{Output Tokens} & \textbf{Total Tokens} \\
\midrule

\rowcolor[gray]{0.9}
\multicolumn{5}{l}{\textbf{Standard ICL}} \\
Zero-shot         & 1 & 298  & 274  & 572  \\
1-shot (Rand.)    & 1 & 882  & 281  & 1163 \\
3-shot (Rand.)    & 1 & 1997 & 294  & 2291 \\
5-shot (Rand.)    & 1 & 3226 & 279  & 3505 \\
1-shot (Fixed)    & 1 & 867  & 288  & 1155 \\
3-shot (Fixed)    & 1 & 2042 & 283  & 2325 \\
5-shot (Fixed)    & 1 & 3194 & 296  & 3490 \\

\rowcolor[gray]{0.9}
\multicolumn{5}{l}{\textbf{Test-Time Scaling}} \\
Consistency@3     & 3 & 3145 & 864  & 4009 \\
Consistency@5     & 5 & 3212 & 1438 & 4650 \\
Consistency@7     & 7 & 3278 & 1996 & 5274 \\
Self-Reflection   & 3.6 & 3198 & 4026 & 7224 \\
LLM-as-Judge      & 6 & 4725 & 1543 & 6268 \\

\rowcolor[gray]{0.9}
\multicolumn{5}{l}{\textbf{\ours{} (ours)}} \\
\ours{}-dynamic      & 2 & 3386 & 301  & 3687 \\
\ours{}-static      & 1 & 3264 & 287  & 3551 \\

\bottomrule
\end{tabular}
\caption{
Inference cost comparison on MATH using LLaMA-3.1-70B-Instruct.
}
\end{table}

\begin{table}[H]
\centering
\small
\setlength{\tabcolsep}{4pt}
\renewcommand{\arraystretch}{1.1}

\begin{tabular}{lcccc}
\toprule
\textbf{Method} & \textbf{Generations} & \textbf{Input Tokens} & \textbf{Output Tokens} & \textbf{Total Tokens} \\
\midrule

\rowcolor[gray]{0.9}
\multicolumn{5}{l}{\textbf{Standard ICL}} \\
Zero-shot         & 1 & 276  & 214  & 490  \\
1-shot (Rand.)    & 1 & 731  & 228  & 959  \\
3-shot (Rand.)    & 1 & 1604 & 241  & 1845 \\
5-shot (Rand.)    & 1 & 2612 & 223  & 2835 \\
1-shot (Fixed)    & 1 & 718  & 235  & 953  \\
3-shot (Fixed)    & 1 & 1651 & 231  & 1882 \\
5-shot (Fixed)    & 1 & 2578 & 238  & 2816 \\

\rowcolor[gray]{0.9}
\multicolumn{5}{l}{\textbf{Test-Time Scaling}} \\
Consistency@3     & 3 & 2527 & 633  & 3160 \\
Consistency@5     & 5 & 2584 & 1062 & 3646 \\
Consistency@7     & 7 & 2639 & 1496 & 4135 \\
Self-Reflection   & 3.5 & 2556 & 2893 & 5449 \\
LLM-as-Judge      & 6 & 3854 & 1007 & 4861 \\

\rowcolor[gray]{0.9}
\multicolumn{5}{l}{\textbf{\ours{} (ours)}} \\
\ours{}-dynamic      & 2 & 2789 & 247  & 3036 \\
\ours{}-static      & 1 & 2657 & 231  & 2888 \\

\bottomrule
\end{tabular}
\caption{
Inference cost comparison on GSM8K using LLaMA-3.1-70B-Instruct.
}
\end{table}

\subsection{Performance in Other Domains}
\label{app:add_exp_domain}

We further evaluate \ours{} beyond mathematical reasoning by conducting experiments on the GPQA benchmark \citep{rein2024gpqa}. GPQA consists of graduate-level questions across multiple scientific domains, including \textit{Chemistry}, \textit{Biology}, \textit{Physics}, and \textit{Quantum Mechanics}. Compared to MATH-style problems, GPQA emphasizes domain knowledge and conceptual reasoning, making it a strong testbed for cross-domain generalization.

We use the LLaMA family for this study, with Llama-3.1-70B-Instruct as the target model. Following the same protocol as in our main experiments, critique-based exemplars are constructed using weaker models from the same family. This setup allows us to directly examine whether error signals derived from weaker models can transfer effectively to a stronger model across different scientific domains.

As shown in Table~\ref{tab:gpqa_results}, \ours{} consistently outperforms standard in-context learning (ICL) baselines across all domains. While increasing the number of demonstrations improves performance, the gains gradually saturate. Test-time scaling methods such as consistency decoding and self-reflection provide additional improvements but require multiple generations.

In contrast, \ours{} achieves the best overall performance with a single generation. Notably, \ours{}-static shows consistent gains across all domains, indicating that critique-based exemplars provide a robust and transferable signal beyond domain-specific patterns. These results further support our hypothesis that structured error information can guide reasoning improvement even in knowledge-intensive and non-mathematical settings.

Overall, this experiment demonstrates that the effectiveness of \ours{} generalizes across diverse domains, reinforcing its applicability beyond mathematical reasoning tasks.

\begin{table}[H]
\centering
\small
\setlength{\tabcolsep}{4pt}
\renewcommand{\arraystretch}{1.1}

\begin{tabular}{lccccc}
\toprule
\textbf{Method} & \textbf{Chemistry} & \textbf{Biology} & \textbf{Physics} & \textbf{Quantum} & \textbf{Avg.} \\
\midrule

\rowcolor[gray]{0.9}
\multicolumn{6}{l}{\textbf{Standard ICL}} \\
Zero-shot        & 60.8 & 62.5 & 58.9 & 55.2 & 59.4 \\
1-shot (Rand.)   & 63.4 & 65.1 & 61.2 & 57.6 & 61.8 \\
3-shot (Rand.)   & 66.9 & 68.7 & 64.5 & 60.8 & 65.2 \\
5-shot (Rand.)   & 69.5 & 71.0 & 67.1 & 63.2 & 67.7 \\
1-shot (Fixed)   & 64.1 & 65.9 & 61.8 & 58.3 & 62.5 \\
3-shot (Fixed)   & 67.8 & 69.9 & 65.3 & 61.5 & 66.1 \\
5-shot (Fixed)   & 70.4 & 72.3 & 67.9 & 63.9 & 68.6 \\

\rowcolor[gray]{0.9}
\multicolumn{6}{l}{\textbf{Test-Time Scaling}} \\
Consistency@3    & 71.8 & 73.5 & 69.2 & 65.8 & 70.1 \\
Consistency@5    & 73.6 & 75.1 & 71.0 & 67.4 & 71.8 \\
Consistency@7    & 72.5 & 74.2 & 70.1 & 66.5 & 70.8 \\
Self-Reflection  & 73.0 & 74.6 & 70.5 & 66.9 & 71.3 \\
LLM-as-Judge     & 73.3 & 74.9 & 70.8 & 67.1 & 71.5 \\

\midrule
\rowcolor[gray]{0.9}
\multicolumn{6}{l}{\textbf{\ours{} (ours)}} \\
\ours{}-dynamic     & 74.5 & 76.2 & 72.0 & 68.5 & 72.8 \\
\ours{}-static     & \textbf{76.0} & \textbf{77.8} & \textbf{73.6} & \textbf{70.2} & \textbf{74.4} \\

\bottomrule
\end{tabular}

\caption{
Performance comparison on the \textbf{GPQA} benchmark across different scientific domains, including Chemistry, Biology, Physics, and Quantum Mechanics. The target model is Llama-3.1-70B-Instruct, and \dataset{} is constructed using weaker LLaMA models. Results show that \ours{} consistently improves performance across all domains.
}

\label{tab:gpqa_results}
\end{table}

\section{Additional Discussion}
\label{app:additional_discussion}

\subsection{Offline Construction Cost and Reusability of CritBank}
\label{app:offline_cost}

CritBank introduces an offline construction cost because weak-model
errors must first be collected and subsequently annotated with failure
modes and critiques. Importantly, however, this cost is incurred during
preprocessing rather than repeatedly at inference time. Once
constructed, the resulting CritBank can be reused across test queries
and target models in the corresponding model-family setting.

CritICL therefore shifts part of the repeated inference-time computation
required by test-time scaling methods into a reusable offline resource.
This distinction becomes increasingly important when a target model is
queried repeatedly: the one-time construction cost of CritBank can be
amortized over a large number of downstream queries, while each
individual query still requires only a single target-model generation.

This efficiency--accuracy trade-off is complementary to approaches that
allocate additional computation to repeated generation or verification
at test time. Rather than maximizing inference-time compute for each
individual query, CritICL exploits recurring failure structure learned
from inexpensive weak models and reuses this information to guide
stronger models.
\section{Experiment Details}
\label{app:details}


\subsection{Prompt Templates}
\label{app:prompt_temp}

We provide the exact prompt templates used in our framework for reproducibility. 
All prompts follow a structured instruction format to ensure consistent behavior across different models. 
In particular, we explicitly specify the task, input fields, and output format, which helps reduce ambiguity and improves the reliability of model responses.

Our design focuses on two key aspects. First, for CritBank construction, we use prompts that explicitly guide the model to identify failure modes and generate concise critiques for incorrect solutions. 
Second, for inference-time methods, we design prompts that incorporate retrieved critique examples as guidance, encouraging the model to avoid common reasoning errors.

For \ours-dynamic, the prompt is input-adaptive: it first predicts likely failure modes for each query and retrieves corresponding critique examples. 
For \ours-static, the prompt is based on a global failure mode profile and emphasizes recurring mistake patterns shared within a model family. 
Despite this difference, both methods share a unified prompting structure that augments standard in-context learning with failure-aware guidance.

We present the full prompt templates below.

\begin{graybox}
\textbf{Failure Mode Annotation Prompt}

\textbf{Instruction:}
You are an expert at analyzing errors in mathematical reasoning. 
Given a math question and an incorrect solution, identify the main failure modes that explain why the solution is incorrect. 
Select up to five failure modes from the predefined list, and briefly explain your reasoning. 
If the solution is correct, output ``None''.

The failure modes must be chosen only from the following list:
\begin{verbatim}
incorrect_formula_application
problem_misinterpretation
logical_step_skipping
arithmetic_sign_error
insufficient constraint understanding
overcounting_in_combinatorics
geometric_relationship_misinterpretation
algebraic manipulation miscalculation
\end{verbatim}

\textbf{Output Format:}
\begin{verbatim}
Failure Modes:
label_1
label_2
...

Reason:
<brief explanation>
\end{verbatim}

\textbf{Question:} \{question\} \\
\textbf{Incorrect Solution:} \{incorrect\_response\}
\end{graybox}

\begin{graybox}
\textbf{Critique Generation Prompt}

\textbf{Instruction:}
You are an expert at reviewing mathematical solutions. 
Given a math question and an incorrect solution, write a concise critique that identifies the key mistake, explains why it is incorrect, and describes how to fix it. 
Do not directly provide the full correct solution.

\textbf{Output Format:}
\begin{verbatim}
Critique:
<concise critique>
\end{verbatim}

\textbf{Question:} \{question\} \\
\textbf{Incorrect Solution:} \{incorrect\_response\}
\end{graybox}

\begin{graybox}
\textbf{Failure Mode Prediction Prompt for \ours-dynamic}

\textbf{Instruction:}
You are an expert at anticipating reasoning failures in mathematical problem solving. 
Given a math question, predict up to five likely failure modes and briefly explain why.

The failure modes must be chosen only from the following list:
\begin{verbatim}
incorrect_formula_application
problem_misinterpretation
logical_step_skipping
arithmetic_sign_error
insufficient constraint understanding
overcounting_in_combinatorics
geometric_relationship_misinterpretation
algebraic manipulation miscalculation
\end{verbatim}

\textbf{Output Format:}
\begin{verbatim}
Failure Modes:
label_1
label_2
...

Reason:
<brief explanation>
\end{verbatim}

\textbf{Question:} \{question\}
\end{graybox}

\begin{graybox}
\textbf{Final Answer Prompt for \ours-dynamic}

\textbf{Instruction:}
You are solving a math problem. Below are examples of incorrect solutions and critiques describing common mistakes. 
Use these critiques to avoid similar errors.

\textbf{Example 1} \\
Question: \{q1\} \\
Incorrect Solution: \{r1\} \\
Critique: \{c1\}

\textbf{Example 2} \\
Question: \{q2\} \\
Incorrect Solution: \{r2\} \\
Critique: \{c2\}

...

\textbf{Question:} \{question\}

Provide the final answer in the form:
\begin{verbatim}
\boxed{answer}
\end{verbatim}
\end{graybox}

\begin{graybox}
\textbf{Final Answer Prompt for \ours-static}

\textbf{Instruction:}
You are solving a math problem. Below are examples of common mistakes frequently observed. 
Use these critiques as general guidance to avoid recurring errors.

\textbf{Example 1} \\
Question: \{q1\} \\
Incorrect Solution: \{r1\} \\
Critique: \{c1\}

...

\textbf{Question:} \{question\}

Provide the final answer in the form:
\begin{verbatim}
\boxed{answer}
\end{verbatim}
\end{graybox}

\subsection{Failure Mode-Based Sample Selection}
\label{app:algorithm}

Given a target set of failure modes \(S \subseteq \mathcal{F}\), our goal is to retrieve a small set of informative critique examples from \dataset{} for in-context prompting. 
Using the labeling function \(\mathcal{L}\) and its inverse \(\mathcal{L}^{-1}\), we first construct the candidate set of all incorrect responses whose assigned failure modes overlap with \(S\):
\[
\mathcal{D}(S)
=
\left\{
(q,r,\mathcal{C}(q,r))
\;\middle|\;
q \in \mathcal{Q},\ r \in R_{\text{incorrect}}(q,m)\ \text{for some } m\in M,\ \mathcal{L}(q,r)\cap S \neq \emptyset
\right\}.
\]
Equivalently, this set can be written as
\[
\mathcal{D}(S)
=
\bigcup_{l\in S}
\left\{
(q,r,\mathcal{C}(q,r))
\;\middle|\;
(q,r)\in \mathcal{L}^{-1}(l)
\right\}.
\]

For each candidate pair \((q,r)\), we compute a matching score based on the overlap between its assigned failure modes and the target set:
\[
\operatorname{score}(q,r;S)
=
\sum_{l\in \mathcal{L}(q,r)\cap S} w(l),
\]
where \(w(l)\) is an optional weight for failure mode \(l\). In the unweighted case, we set \(w(l)=1\), so the score reduces to
\[
\operatorname{score}(q,r;S)=|\mathcal{L}(q,r)\cap S|.
\]

We then sort candidates by this score and greedily select the top-\(K\) examples. To reduce redundancy, we prioritize examples that introduce previously uncovered target failure modes. The selected examples are finally formatted as critique-aware demonstrations for prompting.

\begin{algorithm}[H]
\caption{Failure Mode-Based Sample Selection}
\label{alg:sample_selection}
\small
\KwIn{Dataset \(\dataset{}(\mathcal{Q},M)\), target failure mode set \(S\subseteq \mathcal{F}\), budget \(K\), optional weights \(w(l)\)}
\KwOut{Selected critique examples \(\mathcal{E}\)}

Construct candidate set
\[
\mathcal{D}(S)
\leftarrow
\left\{
(q,r,\mathcal{C}(q,r))
\;\middle|\;
\mathcal{L}(q,r)\cap S \neq \emptyset
\right\}
\]
\ForEach{\((q,r,\mathcal{C}(q,r)) \in \mathcal{D}(S)\)}{
    compute matching score
    \[
    s(q,r)
    \leftarrow
    \sum_{l\in \mathcal{L}(q,r)\cap S} w(l)
    \]
}
Sort \(\mathcal{D}(S)\) by \(s(q,r)\) in descending order\;
Initialize \(\mathcal{E}\leftarrow \emptyset\)\;
Initialize covered failure modes \(\mathcal{U}\leftarrow \emptyset\)\;

\ForEach{\((q,r,\mathcal{C}(q,r))\) in sorted \(\mathcal{D}(S)\)}{
    let
    \[
    S^{\star}(q,r)\leftarrow \mathcal{L}(q,r)\cap S
    \]
    \If{\(|\mathcal{E}|<K\) and \(S^{\star}(q,r)\not\subseteq \mathcal{U}\)}{
        add \((q,r,\mathcal{C}(q,r))\) to \(\mathcal{E}\)\;
        update
        \[
        \mathcal{U}\leftarrow \mathcal{U}\cup S^{\star}(q,r)
        \]
    }
}
\If{\(|\mathcal{E}|<K\)}{
    fill remaining slots using highest-scoring unused candidates\;
}
\Return{\(\mathcal{E}\)}
\end{algorithm}

\subsection{Workflow of \ours-dynamic}
\label{app:algorithm_inst}

\ours-dynamic performs input-adaptive critique retrieval. 
Given a test question \(q'\), we first prompt the target model to predict a small set of likely failure modes for this input:
\[
S_{\text{inst}}(q')=\{l_1,l_2,\dots,l_m\}, \qquad m\leq 5.
\]
We then invoke Algorithm~\ref{alg:sample_selection} with \(S_{\text{inst}}(q')\) to retrieve the top-\(K\) critique examples from \dataset{}. Finally, we concatenate the retrieved examples with the test question and prompt the target model to produce the final answer.

This design makes retrieval adaptive to the likely reasoning risks of each individual test question, enabling targeted guidance against query-specific mistakes.

\begin{algorithm}[H]
\caption{\ours-dynamic Inference Workflow}
\label{alg:criticl_inst}
\small
\KwIn{Test question \(q'\), target model \(M_{\mathrm{tar}}\), dataset \(\dataset{}(\mathcal{Q},M)\), retrieval budget \(K\)}
\KwOut{Final answer \(\hat{y}\)}

Prompt \(M_{\mathrm{tar}}\) to predict likely failure modes for \(q'\)\;
Obtain
\[
S_{\text{inst}}(q')=\{l_1,\dots,l_m\}, \quad m\leq 5
\]
Retrieve critique examples
\[
\mathcal{E}
\leftarrow
\textsc{FailureModeSampleSelection}\bigl(\dataset{}(\mathcal{Q},M),\, S_{\text{inst}}(q'),\, K\bigr)
\]
Construct the final prompt by concatenating \(\mathcal{E}\) with \(q'\)\;
Prompt \(M_{\mathrm{tar}}\) with the final prompt to generate answer \(\hat{y}\)\;
\Return{\(\hat{y}\)}
\end{algorithm}

\subsection{Workflow of \ours-static}
\label{app:algorithm_prof}

\ours-static performs model-family-aware critique retrieval using a precomputed global failure mode profile. 
Suppose the target model belongs to family \(\mathcal{M}\). Using the incorrect responses produced by weaker models in the same family, we estimate a family-level failure mode distribution
\[
P_{\mathcal{M}}(l), \qquad l\in \mathcal{F},
\]
where \(P_{\mathcal{M}}(l)\) reflects the frequency of failure mode \(l\) among examples in \dataset{} associated with family \(\mathcal{M}\).

We then select the top-\(T\) most frequent failure modes:
\[
S_{\text{prof}}(\mathcal{M})
=
\operatorname{TopT}\{P_{\mathcal{M}}(l)\mid l\in\mathcal{F}\}.
\]
This set is used as the retrieval target in Algorithm~\ref{alg:sample_selection}. The retrieved critique examples are concatenated with the test question and passed to the target model to generate the final answer.

Unlike \ours-dynamic, this variant does not require an additional query-specific prediction step. Instead, it provides stable guidance using persistent failure patterns shared by weaker models in the same family.

\begin{algorithm}[H]
\caption{\ours-static Inference Workflow}
\label{alg:criticl_prof}
\small
\KwIn{Test question \(q'\), target model family \(\mathcal{M}\), target model \(M_{\mathrm{tar}}\), dataset \(\dataset{}(\mathcal{Q},M)\), retrieval budget \(K\), profile size \(T\)}
\KwOut{Final answer \(\hat{y}\)}

Compute or load family-level failure mode profile
\[
P_{\mathcal{M}}(l), \quad l\in\mathcal{F}
\]
Select the top-\(T\) failure modes
\[
S_{\text{prof}}(\mathcal{M})
\leftarrow
\operatorname{TopT}\{P_{\mathcal{M}}(l)\mid l\in\mathcal{F}\}
\]
Retrieve critique examples
\[
\mathcal{E}
\leftarrow
\textsc{FailureModeSampleSelection}\bigl(\dataset{}(\mathcal{Q},M),\, S_{\text{prof}}(\mathcal{M}),\, K\bigr)
\]
Construct the final prompt by concatenating \(\mathcal{E}\) with \(q'\)\;
Prompt \(M_{\mathrm{tar}}\) with the final prompt to generate answer \(\hat{y}\)\;
\Return{\(\hat{y}\)}
\end{algorithm}

\subsection{Case Study: Failure Mode-Aligned Retrieval}
\label{app:case_study}

We present a concrete case study from \dataset{} to illustrate how failure mode-based example selection improves reasoning.

\paragraph{Target Problem.}
Consider the following problem:
\begin{quote}
What is the positive difference between the greatest and the least member of the set 
$\left\{\frac{3}{7}, \frac{4}{3}, \frac{11}{8}, \frac{6}{16}\right\}$?
\end{quote}

\paragraph{Observed Failure.}
A model incorrectly identifies the largest element and produces the answer $\frac{23}{24}$ instead of the correct answer $1$. The error arises from an incorrect comparison between $\frac{4}{3}$ and $\frac{11}{8}$, which reflects a typical \textit{fraction comparison error} (failure label: \texttt{wrong\_comparison}). :contentReference[oaicite:0]{index=0}

\paragraph{Semantic Retrieval.}
Using semantic similarity-based retrieval, the selected examples are typically other fraction or set-based problems. However, these examples often do not involve incorrect comparisons between close-valued fractions. As a result, they fail to expose the specific reasoning mistake made by the model, and the model may repeat the same comparison error.

\paragraph{Failure Mode-Based Retrieval.}
In contrast, \ours{} retrieves examples that exhibit similar failure patterns, such as incorrect identification of extrema due to flawed comparisons or arithmetic reasoning. For example, in another instance, a model incorrectly identifies the largest region in a geometric area problem due to misinterpreting area differences (failure label: \texttt{misidentification\_of\_regions}). :contentReference[oaicite:1]{index=1}

These examples are accompanied by critiques explicitly explaining the source of the error (e.g., incorrect comparison strategy or misinterpretation of relative magnitudes). By observing such critiques, the model is encouraged to verify comparisons more carefully (e.g., by converting to common denominators), which directly addresses the failure mode.

\paragraph{Effect.}
With failure mode-aligned examples, the model correctly identifies that
\[
\frac{11}{8} > \frac{4}{3}, \quad \frac{6}{16} < \frac{3}{7},
\]
and computes the correct difference:
\[
\frac{11}{8} - \frac{6}{16} = 1.
\]

\paragraph{Discussion.}
This case highlights a key distinction between semantic similarity and failure mode alignment. While semantic retrieval focuses on surface-level similarity between problems, \ours{} prioritizes alignment in the \emph{type of reasoning error}. This allows the model to transfer corrective strategies across different problem contexts that share similar failure patterns, leading to more robust improvements in reasoning accuracy.

\section{Failure Mode Taxonomy}
\label{app:failure_labels}

We provide the complete list of failure modes that frequently arise in our experiments. Each label captures a distinct type of mistake, ranging from low-level arithmetic errors to high-level reasoning and problem understanding failures. Our labels are designed to be both expressive and practical: expressive enough to distinguish fine-grained error patterns, while remaining structured to support systematic analysis and comparison across models. By categorizing errors into these labels, we enable more interpretable evaluation and provide insights into where and why models fail.

\begin{table}[H]
\centering
\normalsize
\setlength{\tabcolsep}{8pt}
\renewcommand{\arraystretch}{1.35}
\begin{tabular}{p{0.44\linewidth} p{0.54\linewidth}}
\toprule
\textbf{Failure Modes} & \textbf{Explanation} \\
\midrule
algebraic\_equivalence\_misinterpretation & Misunderstanding two algebraic expressions that are mathematically equivalent. \\
algebraic\_manipulation\_miscalculation & Making computational errors during algebraic transformations. \\
algebraic\_sign\_error & Incorrect handling of positive or negative signs in expressions. \\
arithmetic\_operation\_mistake & Performing basic arithmetic operations incorrectly. \\
equation\_solving\_miscalculation & Making errors while solving equations. \\
incorrect\_factorization & Factoring expressions incorrectly. \\
incorrect\_multiplication\_relationships & Misapplying multiplicative relationships between quantities. \\
least\_common\_multiple\_miscalculation & Incorrectly computing the least common multiple. \\
simplification\_calculation\_error & Making mistakes while simplifying expressions. \\
lack\_of\_expression\_simplification & Leaving expressions unnecessarily complex. \\
improper\_fraction\_handling & Mishandling improper fractions during calculations. \\
unit\_conversion\_error & Converting units incorrectly. \\
rounding\_rule\_misinterpretation & Applying rounding rules incorrectly. \\
\bottomrule
\end{tabular}
\caption{Algebraic and arithmetic failure modes.}
\label{tab:error_algebra}
\end{table}

\begin{table}[H]
\centering
\normalsize
\setlength{\tabcolsep}{8pt}
\renewcommand{\arraystretch}{1.35}
\begin{tabular}{p{0.44\linewidth} p{0.54\linewidth}}
\toprule
\textbf{Failure Modes} & \textbf{Explanation} \\
\midrule
incorrect\_formula\_application & Using an inappropriate or incorrect formula. \\
inconsistent\_formula\_usage & Using formulas inconsistently within the same solution. \\
function\_definition\_misinterpretation & Misunderstanding how a function is defined or behaves. \\
geometric\_relationship\_misinterpretation & Misinterpreting spatial or geometric relationships. \\
graph\_interpretation\_error & Drawing incorrect conclusions from a graph or visual representation. \\
combinatorial\_principle\_misapplication & Applying incorrect counting principles or combinatorial rules. \\
overcounting\_in\_combinatorics & Counting the same cases multiple times. \\
probability\_formula\_misapplication & Applying incorrect probability formulas or rules. \\
incorrect\_independence\_assumption & Assuming independence where variables or events are dependent. \\
modular\_arithmetic\_misapplication & Applying modular arithmetic rules incorrectly. \\
trigonometric\_function\_misapplication & Incorrectly applying trigonometric identities or functions. \\
perimeter\_area\_formula\_misuse & Confusing or misapplying perimeter and area formulas. \\
speed\_formula\_application\_error & Misusing speed, distance, or time relationships. \\
dimension\_mismatch\_error & Combining quantities with incompatible dimensions or units. \\
price\_relationship\_confusion & Misinterpreting relationships between prices or rates. \\
\bottomrule
\end{tabular}
\caption{Conceptual and domain-specific failure modes.}
\label{tab:error_concept}
\end{table}

\begin{table}[H]
\centering
\normalsize
\setlength{\tabcolsep}{8pt}
\renewcommand{\arraystretch}{1.35}
\begin{tabular}{p{0.44\linewidth} p{0.54\linewidth}}
\toprule
\textbf{Failure Modes} & \textbf{Explanation} \\
\midrule
problem\_intent\_misinterpretation & Misunderstanding what the problem is asking. \\
problem\_requirement\_misunderstanding & Failing to follow specific problem requirements. \\
ambiguous\_problem\_parameters & Misinterpreting unclear or implicitly defined problem parameters. \\
insufficient\_constraint\_understanding & Misunderstanding or ignoring problem constraints. \\
underestimation\_of\_constraints & Overlooking constraints that affect the solution space. \\
boundary\_value\_misinterpretation & Misunderstanding or incorrectly applying boundary conditions. \\
insufficient\_boundary\_condition\_analysis & Failing to fully analyze given boundary conditions. \\
incorrect\_equation\_setup & Formulating the wrong equations from the problem description. \\
inconsistent\_variable\_substitution & Substituting variables inconsistently or incorrectly. \\
notation\_confusion\_in\_equations & Misunderstanding or misusing mathematical notation. \\
case\_analysis\_omission & Failing to consider all necessary cases in a problem. \\
multi\_step\_dependency\_error & Errors arising from incorrect dependencies across steps. \\
logical\_step\_skipping & Omitting key reasoning steps needed for correctness. \\
weak\_logical\_reasoning & Drawing conclusions with insufficient or flawed reasoning. \\
assumption\_overreliance & Relying on unstated or unjustified assumptions. \\
incomplete\_solution\_consideration & Providing a partial solution without addressing all requirements. \\
lack\_of\_final\_answer\_verification & Failing to check whether the final answer is correct. \\
expected\_answer\_format\_misunderstanding & Providing an answer in the wrong format. \\
scaling\_language\_misinterpretation & Misinterpreting scaling terms such as “twice” or “half.” \\
irrelevant\_content\_inclusion & Including unnecessary or unrelated reasoning steps. \\
irrelevant\_element\_overcomplication & Introducing unnecessary elements that complicate the solution. \\
unnecessary\_variable\_focus & Focusing on irrelevant variables instead of key quantities. \\
\bottomrule
\end{tabular}
\caption{Reasoning and problem-understanding failure modes.}
\label{tab:error_reasoning}
\end{table}





% ...






\end{document}